\documentclass[10pt,twocolumn,letterpaper]{article}

\usepackage[pagenumbers]{cvpr} 

\usepackage{graphicx}
\usepackage{cuted}
\usepackage{multicol}
\usepackage{multirow}
\usepackage{tabularx}
\usepackage{makecell}
\usepackage{amssymb}
\usepackage{pifont}
\usepackage{xcolor,colortbl}
\usepackage{booktabs}
\newcommand{\xmark}{\ding{55}}
\usepackage{comment}
\usepackage{rotating}

\usepackage{amsmath}
\usepackage{cuted}

\definecolor{cvprblue}{rgb}{0.21,0.49,0.74}
\usepackage[pagebackref,breaklinks,colorlinks,allcolors=cvprblue]{hyperref}

\def\paperID{} 
\def\confName{CVPR}
\def\confYear{2026}

\title{Multinex: Lightweight Low-light Image Enhancement via Multi-prior Retinex}

\author{
Alexandru Brateanu$^{1}$, Tingting Mu$^{1}$, Codruta O. Ancuti$^{2}$, Cosmin Ancuti$^{2,3}$\\
\small $^{1}$Department of Computer Science, University of Manchester, Manchester, United Kingdom\\
\small $^{2}$ETcTI, University Politehnica Timisoara, Timisoara, Romania\\
\small $^{3}$West University of Timisoara, Timisoara, Romania\\
}

\begin{document}
\pagenumbering{gobble} 
\maketitle
\begin{strip}
    \centering
     \vspace{-40pt}
     \includegraphics[width=1.0\linewidth,trim={0 0 0 0},clip]{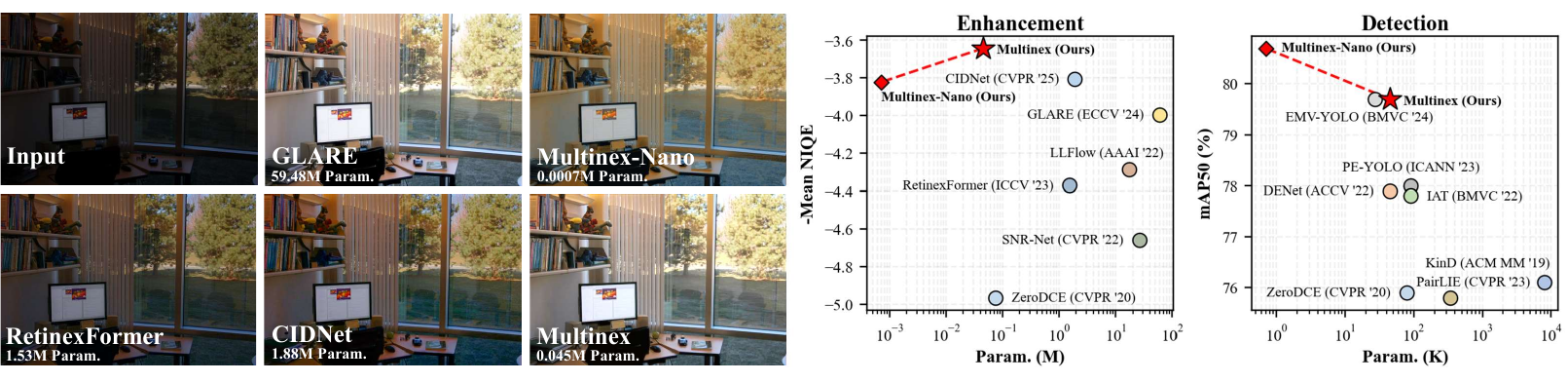}
     \captionof{figure}{Qualitative and quantitative comparison between \textbf{Multinex} variants and recent state-of-the-art methods~\cite{yan2025hvi, GLARE, Retinexformer, LLFlow, SNR-Net, zero_dce, emvyolo, peyolo, denet, iat, KinD, pairlie} on \textit{Enhancement} (no-reference datasets~\cite{MEF, LIME, DICM, NPE}) and \textit{Detection} (ExDark~\cite{Exdark}). Multinex achieves the best perceptual quality, outperforming all compared methods in mean NIQE~\cite{NIQE} while maintaining orders-of-magnitude lower model size, and also attains top mAP50\% in downstream detection.  $-\text{NIQE}$ is used for reporting results so that the top-left region of the plot indicates better performance.}
    \label{fig:teaser}
 \end{strip}

\begin{abstract}
\vspace{-5pt}
Low-light image enhancement (LLIE) aims to restore natural visibility, color fidelity, and structural detail under severe illumination degradation.
State-of-the-art (SOTA)  LLIE techniques often rely on large models  and multi-stage training, limiting  practicality for edge deployment. 
Moreover, their dependence on a single color space  introduces instability and visible exposure or color artifacts.
To address these, we propose \textbf{Multinex}, an ultra-lightweight structured framework that integrates multiple fine-grained representations within a principled Retinex residual formulation. 
It decomposes an image into illumination and color prior stacks derived from distinct analytic representations, and learns to fuse these representations into luminance and reflectance adjustments required to correct exposure.
By prioritizing enhancement over reconstruction and exploiting lightweight neural operations, Multinex   significantly   reduces computational cost, exemplified by its  lightweight (45K parameters) and  nano  (0.7K parameters) versions.
Extensive benchmarks show that all lightweight variants significantly outperform their corresponding lightweight SOTA models, and reach  comparable performance to heavy models. 
Paper page available at \url{https://albrateanu.github.io/multinex}.

\end{abstract}

\section{Introduction}
\label{sec:intro}

Due to poor or  uneven  illumination, many real-world imaging systems produce low-light images.
For instance, night-time surveillance, smart phones, drones and self-driving cars often capture scenes with poor visual quality. 
These images typically suffer from suppressed illumination, poor contrast, and chromatic bias caused by insufficient photon capture and uneven sensor response. 
To address these,  low-light image enhancement (LLIE) aims to restore normal-light appearance by correcting exposure and color  while avoiding noise amplification and structural degradation.

Despite notable progress in LLIE,  challenges remain.
Specifically, coupled modeling of color and brightness  in RGB space  limits enhancement quality, while sensitivity to data-specific color distributions weakens cross-scene generalization.
Early convolution-based deep learning (DL) methods \cite{LLNet,EnGAN,DeepUPE,mbllen}   improve contrast but offer limited physical or perceptual explanation. 
Retinex-driven methods \cite{RetinexNet,Retinexformer}  reintroduce  physically grounded priors through illumination-reflectance decomposition, yet still  operate largely in RGB space where luminance and chrominance remain coupled,  undermining  Retinex-style decoupling.
To address this, some approaches  reformulate the color domain by using YCbCr or YUV space \cite{Bread,brateanu2025lyt}, HSV space \cite{zhang2021better, zhou2023low}, and learnable color spaces such as HVI \cite{yan2025hvi}.
%
Although improvements have been made, issues such as partial luminance-color coupling, red discontinuity, black-plane noise, and training instability remain, leading to  degraded enhancement quality across challenging scenes.


%
Another challenge is that state-of-the-art (SOTA) LLIE techniques  rely on large neural networks with (tens of) millions of parameters,    limiting  real-time edge deployment.
This has motivated research on lightweight LLIE  \cite{pairlie, zero_dce, brateanu2025lyt}.
However, drastically reducing model size (e.g.,  below 1M parameters)   degrades enhancement quality.
To address these challenges, we  improve LLIE  by enabling \textit{stable information decoupling} between illumination and color, with a focus on  \emph{lightweight LLIE}  to mitigate   performance drop under extreme parameter reduction. 


%
To enable low-cost, effective LLIE, we propose  a compact and physically grounded framework, named  Multi-prior Retinex (\textbf{Multinex}).
It integrates   analytic decomposition guided by  Retinex theory,  multi-view representation priors derived from classical  color theory, and lightweight  learnable fusion.
Specifically,  Multinex  introduces a  residual decomposition, which, unlike the  implicit Retinex decomposition, models only the required image adjustment, emphasizing enhancement over reconstruction.
It preserves  structure,  reduces dataset bias, and  lowers computational cost.
Instead of learning a color-space transformation, Multinex analytically extracts complementary representation priors for lightness and chromatic structure.
This  eases  representation learning under tight parameter budget. 
It then constructs lightweight learnable modules to  fuse the luminance and chrominance representations, with     efficient  neural operation design and careful parameter allocation.
This enables extreme   model size reduction.
Extensive experiments show that Multinex  achieves 
color-consistent and real-time enhancement supported by physical explanation, achieving strong performance in both lightweight (45K parameters) and nano (0.7K parameters) configurations.
Our main contributions are summarized as follows:
\begin{itemize}
    \item Identify luminance-color decoupling as a bottleneck in LLIE, and introduce analytic representation priors \emph{illumination and reflectance guidance stacks} to  separate brightness and chromatic cues and enable decoupled learning.
    %
    \item Propose Retinex-guided residual formulation, termed \emph{enhancement delta},  emphasizing enhancement over reconstruction and yielding more effective correction structure.
    %
    \item Design a dual-branch lightweight   fusion network to refine the guidance stacks, achieving extreme parameter efficiency, e.g.,   92\% median parameter reduction and 89\% median FLOP reduction.
    \item Empirically demonstrate that Multinex consistently outperforms SOTA on 5 metrics and 7 benchmark datasets under extreme model compression.
    %
    %

\end{itemize}


\vspace{-0.1cm}
\section{Related Work}
\label{sec:related_work}


\subsection{Deep Learning for LLIE }
\label{sec:deep_learning}

LLIE  techniques have evolved from  classical image processing (e.g., gamma correction and histogram equalization)   to deep learning (DL)  that has   dominated the SOTA since around 2018.
Early DL methods   build on end-to-end convolutional neural networks (CNNs) \cite{LLNet,EnGAN,DeepUPE,mbllen}.
They learn how to brighten under-exposed scenes directly from paired data, e.g., low-light vs. normal-light images, achieving significant gains in contrast recovery and noise suppression. 
However, their limited receptive fields can hinder long-range coherence  and   lead  to localized exposure imbalance.  %
Transformer-based architectures address this by modeling global dependency    through   self-attention mechanisms,  achieving SOTA perceptual quality.
To mitigate the quadratic cost of  attention calculation,   windowed or hierarchical mechanisms  are developed \cite{Retinexformer,LLFormer,Restormer}.  
More recently, diffusion  models \cite{Ho2020Denosing,yi2023diff,hou2024global} are applied  to tackle LLIE through a generative perspective,  simulating the gradual denoising and exposure correction.
They deliver SOTA realism  but with slow  inference.

Despite the  success, DL-based LLIE methods demand large model size and   suffer  from high training cost. 
For instance, most transformer-based LLIE models contain  millions of parameters \cite{Retinexformer, SNR-Net, yan2025hvi}, while diffusion-based ones require tens of millions of parameters and tens to hundreds of giga floating-point operations per second (GFLOPS) \cite{hou2024global, quadprior}.
This limits their practicality in real-time and resource-limited environment and edge scenarios, e.g., surveillance systems, smart phones and drones, etc.

%
\subsection{Retinex and Color Frameworks }
\label{sec:retinex_color}


Retinex theory has influenced both classical and DL-based LLIE techniques, remaining a cornerstone for physically grounded image enhancement. 
By decomposing an image   into   illumination and reflectance, it decouples brightness correction and detail preservation.
Classical methods  \cite{SSR,MSRCR} estimate  illumination  through hand-crafted priors, whereas DL methods \cite{RetinexNet,KinD,DeepUPE,Retinexformer} learn illumination and/or reflectance maps (and/or related representations)  by end-to-end neural networks  trained on dedicated loss functions. 
For instance, Retinex-Net~\cite{RetinexNet} explicitly estimates the illumination and reflectance  maps by two CNNs.
KinD \cite{KinD} models the two maps from Retinex decomposition, along with additional reflectance  restoration and illumination adjustment, using three CNNs.
RetinexFormer \cite{Retinexformer} derives from Retinex theory an LLIE approximation as a composition of an illumination estimator and a  corruption restorer implemented with a CNN and a transformer.

%


Recent works extend the Retinex principle to color spaces that explicitly separate luminance and chrominance,  improving  enhancement stability. 
Conversions such as YCbCr~\cite{Bread}, YUV~\cite{brateanu2025lyt}, and HSV~\cite{zhang2021better, zhou2023low} partially disentangle brightness from hue, reducing channel correlation and hue distortion. 
Nevertheless, each space introduces new artifacts, such as hue discontinuity in HSV or residual entanglement, restricting universal applicability. 
Recent HVI-based LLIE~\cite{yan2025hvi} introduces learnable mapping functions to better decouple illumination and chromaticity, but these   are highly data-dependent and can be unstable during training and inference.  
In contrast, our work   takes advantage of both Retinex and color principles to improve  efficiency of lightweight LLIE models.


\subsection{Lightweight  LLIE}
\label{sec:efficiency}

As noted,  SOTA LLIE techniques rely on large neural networks with (tens of) millions of parameters  \cite{Retinexformer,LLFlow,GLARE,quadprior}, requiring substantial computation thus hindering  edge  deployment. 
Efficiency has  therefore  become a key focus in  LLIE research, driving growing interest in  lightweight development.
Designing lightweight neural networks demands careful choices of architecture and parameter allocation, to achieve significant parameter reduction without compromising   effectiveness. 
For instance, PairLIE \cite{pairlie}  reduces its network size to 330K trainable   parameters, by isolating its illumination learning within a lightweight Retinex formulation.
ZeroDCE~\cite{zero_dce}  reduces  its network size  to less than 80K  parameters, by learning pixel-wise curve mappings through handcrafted exposure losses.
LYT-Net~\cite{brateanu2025lyt}    constructs an efficient transformer of 45K parameters in the YUV color space, by leveraging   channel decoupling as an alternative to Retinex decomposition.
Typically, lightweight architectures  trade off   enhancement  quality with noticeable  degradation in lightweight (under 1M parameters) and micro (under 10K parameters) models.
In this work,  we aim to reduce such performance gap by   a simpler yet more robust lightweight model design.



%

\section{Proposed Method: Multinex}
\label{sec:our_approach}

\begin{figure*}
    \centering
    \includegraphics[width=1.0\linewidth]{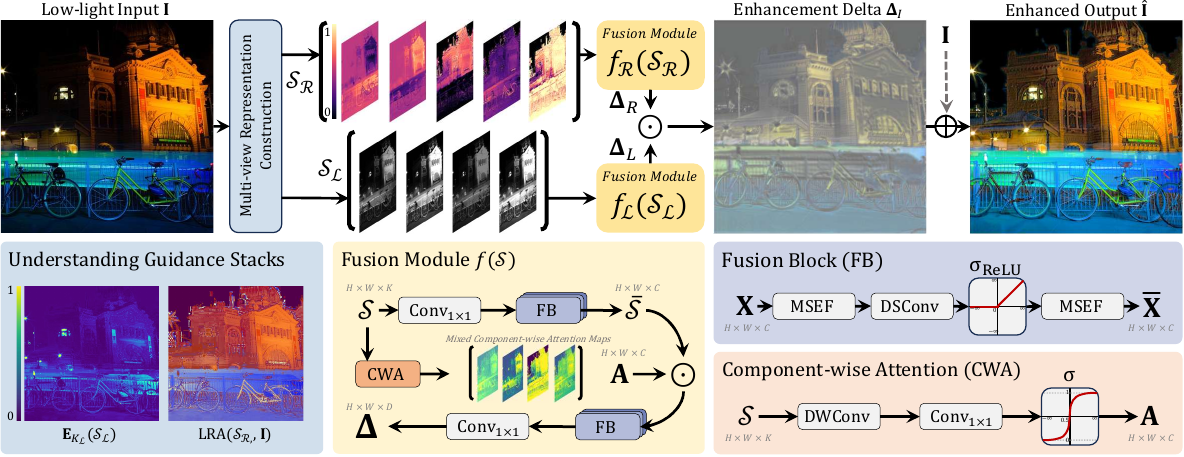}
    \caption{Illustration of Multinex Architecture. Fusion modules $f_\mathcal{L}$ and $f_\mathcal{R}$ use illumination and reflectance guidance stacks $\mathcal{S}_\mathcal{L}$ and $\mathcal{S}_\mathcal{R}$ to produce color and luminance correction terms $\boldsymbol{\Delta}_L$ and $\boldsymbol{\Delta}_R$. Retinex-like fusion then yields the Enhancement Delta $\boldsymbol{\Delta}_I$.}
    \vspace{-10pt}
    \label{fig:framework}
\end{figure*}

The design motivation of Multinex is to exploit, through a \emph{lightweight  model}, as much information and cues embedded within low-light images, aiming at  effective low-cost image enhancement.
When correcting illumination deficiencies in low-light images, existing approaches mostly rely on single color space, e.g. RGB, YUV~\cite{brateanu2025lyt}, HVI~\cite{yan2025hvi}, overlooking useful complementary cues present in the input.
We propose a structured solution that learns from multiple fine-grained illumination and chrominance representations  complementing each other, and fuses these representations into a principled Retinex-based residual formulation to enhance images, guided by classical color vision theory and underpinned by  modern neural building blocks (see   Multinex architecture   in Fig. \ref{fig:framework}). 
%


\subsection{Additive Retinex-based Enhancement Delta}


Denote a low-light RGB image  by a 3-dimensional (3D) tensor $\mathbf{I}\in[0,1]^{H\times W\times 3}$.
%
The goal of LLIE is to compute an enhanced image $\hat{\mathbf{I}}\in [0,1] ^{H\times W\times 3}$ from the input $\mathbf{I}$.
We refer to the $i$-th matrix slice along the  depth of a 3D tensor  $\mathbf{X}\in \mathbb{R}^{H\times W\times C}$ by $\mathbf{X}_i \in \mathbb{R}^{H\times W}$ with $i\in\{1,2,\ldots, C\}$.
%
%
%
%
%
When applying the basic Retinex theory to LLIE,  the enhanced image is expressed as $\hat{\mathbf{I}} =  \mathbf{L}  \odot  \mathbf{R} $, where $ \mathbf{L} $ and $ \mathbf{R} $ are the enhanced luminance and reflectance maps,  and $\odot$ denotes the Hadamard product. 
%
%
%


In practice, such a direct application of Retinex decomposition for the purpose of reconstruction can be inefficient and lack robustness in practice.
Because illumination and color cues remain entangled in RGB, it is challenging to recover a faithful decomposition  under low exposure.  
%
%
To address this, we propose to treat the Retinex decomposition as a structural prior, rather than a reconstruction output. 
Instead of predicting  the two terms $ \mathbf{L} $ and $ \mathbf{R} $ so that their multiplication can recover $\hat{\mathbf{I}}$, we estimate  an additive correction field  that adjusts the input to produce a well-lit output, termed as the \emph{enhancement delta} $\boldsymbol\Delta_I$.
We further factor $\boldsymbol\Delta_I$ into two adjustment fields $\boldsymbol\Delta_L$ and $\boldsymbol\Delta_R$, which are responsible for luminance and reflectance corrections, respectively, aligning with Retinex as a structural prior.
This results in  
\begin{equation}
    \hat{\mathbf{I}} = \mathbf{I} + \boldsymbol\Delta_I =    \mathbf{I} + \boldsymbol\Delta_{L} \odot  \boldsymbol\Delta_{R}. 
    \label{eq:initial_formulation}
\end{equation}
We directly model the  luminance  correction $\boldsymbol\Delta_{L} $ shared among the three RGB channels  by an illumination neural network  $ f_{\mathcal{L}}(\mathbf{I}, \theta_\mathcal{L}) \in \mathbb{R}^{H\times W\times 1}$, while separately model the reflectance correction $\boldsymbol\Delta_{R}$ by  a reflectance  neural network  $ f_{\mathcal{R}}(\mathbf{I}, \theta_\mathcal{R}) \in \mathbb{R}^{H\times W\times 3}$, where \(\theta_\mathcal{L}\) and \(\theta_\mathcal{R}\) denote the network weights. 
%
As a result, each enhanced image slice  $\hat{\mathbf{I}}_i \in [0,1] ^{H\times W}$  is computed by 
\begin{equation}
\hat{\mathbf{I}}_i =   \mathbf{I}_i+  f_{\mathcal{L}}(\mathbf{I}, \theta_\mathcal{L}) \odot f_{\mathcal{R}_i}(\mathbf{I}, \theta_\mathcal{R}),
\label{eq:final_fusion}
\end{equation}
for $i\in\{1,2,3\}$, where  $f_{\mathcal{R}_i}$ denotes the $i$-th  matrix slice output by the network $f_{\mathcal{R}}$.
There is no restrictive activation applied  to the output of $f_{\mathcal{L}}$ and $f_{\mathcal{R}}$  to encourage flexible correction.   
Since $\mathbf{I}$  contains intrinsic texture information, we  let  $f_{\mathcal{R}}$  focus on   extracting  color information that complements  $f_{\mathcal{L}}$ for extracting  lighting information.

The residual formulation in Eq. (\ref{eq:final_fusion}) results in more  effective enhancement at lower cost  than the direct Retinex decomposition. 
Define $\boldsymbol\Delta_{L} = f_{\mathcal{L}}(\mathbf{I}, \theta_\mathcal{L})$ as   \emph{Multinex Luminance} that encodes per-pixel lightness adjustments, and   $\boldsymbol\Delta_{R} = f_{\mathcal{R}}(\mathbf{I}, \theta_\mathcal{R})$  as \emph{Multinex Reflectance} that captures color corrections. 
Using $\boldsymbol\Delta_{L}$ and $\boldsymbol\Delta_{R}$  as adjustment terms prevents the loss of inherent key details and structural information from the original image, and avoids color shifts.
%
%
Under the lightweight constraint,   it  is effective to directly enforce the original image structure through retaining  $\mathbf{I}$, and refine   image content through approximating an additive residual by following   basic Retinex and color principles.  
%



\subsection{Multi-view Representation Construction}

To facilitate lightweight  construction, we  reduce the burden of representation learning by supplying the networks with predefined physical and perceptual cues that are computed analytically.
These are referred to as the \emph{luminance guidance stack}  $\mathcal{S}_\mathcal{L}$ and \emph{reflectance guidance stack} $\mathcal{S}_\mathcal{R}$,  serving as a diverse set of priors that capture brightness, extreme values, and color invariants.
Subsequently, the enhancement  formulation in Eq. (\ref{eq:final_fusion}) is   revised to the following:  
\begin{equation}
\hat{\mathbf{I}}_i = \mathbf{I}_i +f_{\mathcal{L}}(\mathcal{S}_\mathcal{L}(\mathbf{I}), \theta_\mathcal{L}) \odot f_{\mathcal{R}_i}(\mathcal{S}_\mathcal{R}(\mathbf{I}), \theta_\mathcal{R}).
\label{eq:final_fusion_stack}
\end{equation}
%
Each stack contains  $K_\mathcal{L}$ or  $K_\mathcal{R}$ maps computed by formulations for color space conversion and RGB-based descriptors \cite{brateanu2025lyt, yan2025hvi, shi2024zero}, resulting in $\mathcal{S}_\mathcal{L}(\mathbf{I}) \in [0,1] ^{H\times W\times K_\mathcal{L}}$ and   $\mathcal{S}_\mathcal{R}(\mathbf{I}) \in [0,1] ^{H\times W\times K_\mathcal{R}}$. 
When selecting ways to formulate these luminance and chrominance maps, we attempt to widen  the covered range of physical and perceptual cues, while reduce   information overlapping between the maps.
%
%
Below  we explain the selected feature maps, for which   more studies are in Section \ref{sec:supp_guidance_study} of  supplementary material.

 
\noindent\textbf{Luminance Guidance Stack.} 
A total of $K_\mathcal{L}=4$ illumination feature maps (also referred to as luminance channels) are included, each offering a distinct perceptual or physical interpretation of scene brightness.
 This enables the model to reason about exposure and contrast from complementary cues. 
Specifically, we use
\begin{equation}
    \mathcal{S}_\mathcal{L}(\mathbf{I}) = \left[ \mathbf{Y}_\text{Rec.709}, \mathbf{Y}_\text{vmax}, \mathbf{Y}_\text{lightness}, \mathbf{Y}_{\text{L}_2}\right],
    \label{eq:illum_stack}
\end{equation}

where    the concatenation  $[\cdots]$   integrates matrix slices into a tensor. 
Each selected feature map is computed by 
\begin{align}
\label{Y_rec709}
  &  \mathbf{Y}_{\text{Rec.709}} =  0.2126\,\mathbf{I}_R + 0.7152\,\mathbf{I}_G + 0.0722\,\mathbf{I}_B, \\
  &   \mathbf{Y}_{\text{vmax}} =  \max(\mathbf{I}_R, \mathbf{I}_G, \mathbf{I}_B),\\
  &   \mathbf{Y}_{\text{lightness}} =   0.5\big(\max(\mathbf{I}_R, \mathbf{I}_G, \mathbf{I}_B) + \min(\mathbf{I}_R, \mathbf{I}_G, \mathbf{I}_B)\big), \\
  &    \mathbf{Y}_{\text{L}_2} = \sqrt{\mathbf{I}_R^2 + \mathbf{I}_G^2 + \mathbf{I}_B^2 + \varepsilon},
\end{align}
where  $\varepsilon=1e-6$ is included to avoid  zero output for numerical stability. The   operations  $(\cdot)^2$ and $\sqrt{(\cdot)}$ used by computing $\mathbf{Y}_{\text{L}_2}$ are applied element-wise.
Here we  refer to  the   matrix slices of an image   by its channel name, e.g., RGB, instead of the depth index, e.g., $\mathbf{I}_i$, for better clarity.

The feature map $\mathbf{Y}_{\text{Rec.709}}\in [0,1]^{H\times W}$ follows the ITU-R BT.709 standard \cite{ITU709}  and reflects the non-uniform sensitivity of  human eyes towards green,  serving as the most faithful approximation of visual brightness.  
  $\mathbf{Y}_{\text{vmax}}\in [0,1]^{H\times W}$  captures the brightest response among channels, acting as a proxy for highlight energy.
 $\mathbf{Y}_{\text{lightness}}\in [0,1]^{H\times W}$  approximates perceptual lightness in the HSL color space  \cite{HSL}, introducing slight contrast regularization.  
Finally,   $\mathbf{Y}_{\text{L}_2}\in [0,1]^{H\times W}$ represents the overall energy of the RGB vectors, encoding pixel magnitude of the input image.


\noindent\textbf{Reflectance Guidance Stack.} 
A total of $K_\mathcal{R}=5$ color feature maps (also referred to as chrominance channels) are used,  capturing the color-difference and hue information that often degrades under low illumination, resulting in 
\begin{equation}
    \mathcal{S}_\mathcal{R}(\mathbf{I})  = \left[\mathbf{C}_b, \mathbf{C}_r, \mathbf{r}, \mathbf{g}, \mathbf{S}\right].
    \label{eq:reflectance_stack}
\end{equation}
Each feature map enables a different and decoupled view of the input color, computed by  
\begin{align}
     \mathbf{C}_b  =\; & -0.168736\,\mathbf{I}_R - 0.331264\,\mathbf{I}_G + 0.5 \,\mathbf{I}_B,\\
     \mathbf{C}_r  = \; &  0.5 \,\mathbf{I}_R - 0.418688\,\mathbf{I}_G - 0.081312\,\mathbf{I}_B, \\
     \mathbf{r} =\; & \frac{\mathbf{I}_R}{\mathbf{I}_R + \mathbf{I}_G + \mathbf{I}_B + \varepsilon},  \\
     \mathbf{g} = \; &\frac{\mathbf{I}_G}{\mathbf{I}_R + \mathbf{I}_G + \mathbf{I}_B + \varepsilon}, \\
      \mathbf{S} = \; &\frac{\max(\mathbf{I}_R, \mathbf{I}_G, \mathbf{I}_B) - \min(\mathbf{I}_R, \mathbf{I}_G, \mathbf{I}_B)}{\max(\mathbf{I}_R, \mathbf{I}_G, \mathbf{I}_B) + \varepsilon}.
\end{align}
where $\varepsilon=1e-6$ is included to    avoid   division by zero, ensuring numerical stability.

Three sets of color maps are used. 
The  reflect-color-plane set consists of two feature maps  $\mathbf{C}_b, \mathbf{C}_r \in [0,1]^{H\times W}$,   representing the blue-difference and red-difference  relative to the BT.709 luminance \cite{ITU709}.
They effectively isolate the color information from the global intensity.
The second set contains $\mathbf{r}, \mathbf{g}\in [0,1]^{H\times W} $,  encoding chromaticity ratios that represent the red ($\mathbf{I}_R$) and green ($\mathbf{I}_G$) channels normalized by the total pixel intensity.
They are invariant to absolute illumination scale, preserving color relationship.  
Finally, the  map  $\mathbf{S} \in [0,1]^{H\times W}$  reflects how far a pixel lies from the gray axis, which is the saturation measure for quantifying the vividness of color relative to the brightness.

\subsection{Visualizing the Physics of Guidance Stacks}
\label{visual_methods}

To  better understand the physical meaning of the used guidance stacks $\mathcal{S}_{\mathcal{L}}$ and $\mathcal{S}_{\mathcal{R}}$, we introduce ways of   visualizing and analyzing each stack for a given  image.
We demonstrate their use for   validating prior choices  in Section \ref{sec:supp_guidance_study} of supplementary material.

\noindent\textbf{Luminance $\boldsymbol{\mathcal{S}_{\mathcal{L}}}$.}
Each feature map  in the luminance stack, i.e., $\mathcal{S}_{\mathcal{L}_c}$ for $c \in\{ 1,2,\ldots,K_\mathcal{L}\}$, is a grayscale  luminance descriptor with clear interpretability. 
To understand how these descriptors respond differently across the image, we introduce  \emph{Descriptor Importance Analysis} (DIA) by examining complementarity in local structure and global shading.
To extract local structural information, we  employ a simple edge detection method of maximum spatial gradient, producing a gradient map.
To measure global shading expressivity, we   compute principal component analysis (PCA) guided orthogonal energy for each pixel across  different feature maps, resulting in an energy map.  
We then define the gradient and energy importance for a feature map by examining the reduction in the corresponding maps after removing that feature map from the stack. 
Fig.~\ref{fig:framework} visualizes the energy map, where higher activations indicate stronger complementarity within the priors in $\mathcal{S}_{\mathcal{L}}$. 
Section \ref{supp_illum_importance} of  supplementary material details   the importance analysis.

\noindent\textbf{Reflectance  $\boldsymbol{\mathcal{S}_{\mathcal{R}}}$.}
The chromatic feature maps are not directly interpretable because color is distributed across the descriptors. 
Therefore, we introduce a way to visualize   how well these descriptors jointly explain image color through a \textit{Linear Reconstruction Analysis} (LRA), which applies PCA and ridge regression to reconstruct a target image.
%
We explain  its  mathematical formulation in Section \ref{sec:supp_lra} of supplementary material.
%
The reconstruction example in Fig.~\ref{fig:framework} shows $\mathcal{S}_{\mathcal{R}}$ jointly captures color structure, supporting its use for reflectance modeling.

\subsection{Lightweight Neural Network Architecture}

The  luminance and reflectance guidance stacks $\mathcal{S}_\mathcal{L}$ and $\mathcal{S}_\mathcal{R}$ aggregate multiple brightness and color-based priors,   going beyond single color-space representations. 
We design   illumination  and reflectance  networks, i.e., $f_\mathcal{L}$ and $f_\mathcal{R}$   in Eq. (\ref{eq:final_fusion_stack}),  to effectively leverage these these complementary cues.
This amounts  to extracting high-quality information  from   multi-view representations, i.e., the  feature maps in $\mathcal{S}_\mathcal{L}$ for $f_\mathcal{L}$ and in $\mathcal{S}_\mathcal{R}$ for $f_\mathcal{R}$, to improve  enhancement quality.
We adopt the same architecture  for both networks, but with separate  weights and independent operations.
For  convenience, we describe the  architecture using a generic  function   $f(\mathcal{S}): \mathbb{R}^{H\times W\times K} \rightarrow \mathbb{R}^{H\times W\times D}$, where $K=4$, $D=1$ for   $f_\mathcal{L}$, while $K=5$, $D=3$ for   $f_\mathcal{R}$.
We refer to $f(\mathcal{S})$ as a \emph{fusion module} as it fuses the feature maps in $\mathcal{S}$.

\subsubsection{Fusion Module Design}

To construct the fusion module, we propose two lightweight operations,  including a \emph{fusion block}  $\text{FB}: \mathbb{R}^{H\times W\times C} \rightarrow \mathbb{R}^{H\times W\times C}$ responsible for information  refinement and a    \emph{component-wise attention mechanism} $\text{CWA}: \mathbb{R}^{H\times W\times C} \rightarrow  [0,1]^{H\times W\times C}$ to integrate local and global cues.
The basic neural operations used by FB and CWA (e.g., the convolution variants and MSEF \cite{brateanu2025lyt}) are explained in Section \ref{app:multinex_detail} of supplementary material.
%

\noindent\textbf{Fusion Block (FB).}
%
Given the input feature  $\mathbf{X}  \in \mathbb{R}^{H\times W\times C}$, the information refinement block FB is a composite function returning the following output
\begin{equation}
\label{eq:fb}
\bar{\mathbf{X}}  =\mathrm{MSEF} \circ\sigma_{\text{ReLU}} \circ \mathrm{DSConv} \circ \mathrm{MSEF} (\mathbf{X} ),
\end{equation}
where a  $3{\times}3$  filter  is used by $\mathrm{DSConv}$.
The first MSEF adaptively calibrates   each channel  using global context.
$\mathrm{DSConv}$  with ReLU activation introduces lightweight spatial filtering and smooth gating. 
The second MSEF then re-evaluates the enhanced activations, reinforcing cross-channel consistency. 
Overall,   FB   refines  and re-weights  the input by a sequential application of local filtering and adaptive channel-wise recalibrating, yielding efficient, gradient-stable fusion across both local and global scales.



\begin{table*}[t]
\centering
\scriptsize
\setlength{\tabcolsep}{3pt}
\renewcommand{\arraystretch}{1.1}
\resizebox{\textwidth}{!}{
\begin{tabular}{l l c c c c c c c c c c c c c c}
\toprule
&  & 
\multicolumn{3}{c}{\textbf{LOLv1 Dataset}} & & 
\multicolumn{3}{c}{\textbf{LOLv2-real Dataset}} & & 
\multicolumn{3}{c}{\textbf{LOLv2-syn Dataset}} & & 
\multicolumn{2}{c}{\textbf{Model Size}} \\
\cmidrule(lr){3-5}\cmidrule(lr){7-9}\cmidrule(lr){11-13}\cmidrule(lr){15-16}
& \multirow{-2.5}{*}{\textbf{Method}} & PSNR$\uparrow$ & SSIM$\uparrow$ & LPIPS$\downarrow$ && PSNR$\uparrow$ & SSIM$\uparrow$ & LPIPS$\downarrow$ && PSNR$\uparrow$ & SSIM$\uparrow$ & LPIPS$\downarrow$ && Param. (M)$\downarrow$ & GFLOPs$\downarrow$ \\
\midrule

& QuadPrior~\cite{quadprior} & 22.85 & 0.800 & 0.201 && 20.59 & 0.811 & 0.202 && 16.11 & 0.758 & 0.144 && 1252.75 & 1103.2 \\
& GLARE~\cite{GLARE} & \textbf{23.55} & \textbf{0.863} & \textbf{0.086} && \textbf{22.51} & \textbf{0.871} & \textbf{0.105} && 18.21 & 0.842 & 0.178 && 59.48 & 508.42 \\
& LLFlow~\cite{LLFlow}  & 21.15 & 0.854 & 0.116 && 17.43 & 0.831 & 0.176 && \textbf{24.81} & 0.919 & 0.067 && 17.42 & 358.4 \\
\multirow{-4}{*}{\rotatebox[origin=c]{90}{\begin{tabular}{c} \textbf{Heavy} \\ ($>10$M) \end{tabular}}}
& GSAD~\cite{hou2024global} & 22.77 & 0.852 & 0.102 && 20.15 & 0.846 & 0.113 && 24.47 & \textbf{0.929} & \textbf{0.051} && 17.36 & 442.02 \\
\midrule

& KinD~\cite{KinD} & 17.65 & 0.775 & 0.207 && 17.54 & 0.669 & 0.375 && 18.32 & 0.796 & 0.252 && 8.02 & 34.99 \\
& Bread~\cite{Bread}  & 22.96 & 0.838 & 0.155 && 20.83 & 0.847 & 0.174 && 17.63 & 0.919 & 0.091 && 2.02 & 19.85 \\
& CIDNet~\cite{yan2025hvi} & {23.81} & \textbf{0.857} & \textbf{0.086} && \textbf{24.11} & \textbf{0.871} & \textbf{0.108} && \textbf{25.71} & \textbf{0.942} & \textbf{0.045} && {1.88} & {7.57} \\
\multirow{-4}{*}{\rotatebox[origin=c]{90}{\begin{tabular}{c} \textbf{Mid-size} \\ ($1$--$10$M) \end{tabular}}}
& RetinexFormer~\cite{Retinexformer}  & \textbf{25.15} & 0.846 & {0.131} && 22.79 & 0.840 & {0.171} && {25.67} & 0.930 & {0.059} && {1.53} & {15.85} \\
\midrule

& RetinexNet~\cite{RetinexNet}  & 16.77 & 0.419 & 0.470 && 16.10 & 0.401 & 0.543 && 17.14 & 0.762 & 0.255 && 0.8400 & 584.47 \\
& PairLIE~\cite{pairlie}  & 18.47 & 0.743 & 0.290 && 19.89 & 0.778 & 0.317 && 19.07 & 0.794 & 0.230 && 0.3300 & 20.81 \\
& ZeroDCE~\cite{zero_dce}  & 14.86 & 0.559 & 0.335 && 16.06 & 0.580 & 0.313 && 17.71 & 0.815 & 0.169 && 0.0790 & 4.83 \\
& LYT-Net~\cite{brateanu2025lyt} & {22.38} & {0.826} & {0.182} && {21.83} & {0.849} & 0.225 && {23.78} & {0.921} & 0.097 && {0.0449} & {3.49} \\ 
\rowcolor{cvprblue!14} \cellcolor{white} 
\multirow{-5}{*}{\rotatebox[origin=c]{90}{\begin{tabular}{c} \cellcolor{white}\textbf{Lightweight} \\ \cellcolor{white}($<1$M) \end{tabular}}}
& \textbf{Multinex} & \textbf{23.19} & \textbf{0.843} & \textbf{0.129} && \textbf{23.04} & \textbf{0.860} & \textbf{0.178} && \textbf{25.04} & \textbf{0.930} & \textbf{0.068} && 0.0446 & 2.50 \\
\midrule

& RUAS~\cite{RUAS} & 16.40 & 0.503 & 0.303 && 15.33 & 0.493 & 0.325 && 13.40 & 0.640 & 0.365 && 0.00343 & 0.83 \\
& RSFNet~\cite{RSFNet}  & 19.39 & \textbf{0.745} & 0.278 && 19.46 & 0.745 & 0.278 && 17.18 & 0.817 & 0.159 && 0.00211 & -- \\
& SCI~\cite{SCI}  & 14.78 & 0.525 & 0.327 && 17.30 & 0.540 & 0.294 && 15.43 & 0.744 & 0.203 && 0.00026 & 0.06 \\ 
\rowcolor{cvprblue!14}\cellcolor{white} 
\multirow{-4}{*}{\rotatebox[origin=c]{90}{\begin{tabular}{c} \cellcolor{white} \textbf{Micro} \\ \cellcolor{white} ($<10$k) \end{tabular}}}
& \textbf{Multinex-Nano} & \textbf{19.42} & {0.742} & \textbf{0.276} && \textbf{19.66} & \textbf{0.784} &\textbf{0.266} && \textbf{21.05} & \textbf{0.882} & \textbf{0.143} && {0.00069} & 0.04 \\
\bottomrule
\end{tabular}
}
\vspace{-5pt}
\caption{Results on LOLv1~\cite{RetinexNet}, and LOLv2~\cite{Sparse} datasets (real and synthetic) across four model groups of parameter scales. Best performance  in each group is highlighted in \textbf{bold}.}
\vspace{-10pt}
\label{tab:llie_vertical_grouped}
\end{table*}

\noindent\textbf{Component-wise Attention (CWA).}
We design a lightweight attention mechanism that computes soft attention scores to selectively weight components derived from local regions of the input feature map.
Specifically, given the input $\mathbf{X} \in \mathbb{R}^{H\times W\times C}$, we construct a channel-aware attention map $\mathbf{A} \in [0,1]^{ H\times W\times C^\prime}$  by 
\begin{equation}
%
%
\mathbf{A}  =  \sigma   \circ \mathrm{Conv}_{1\times1} \circ \mathrm{DWConv}(\mathbf{X} ). 
\end{equation}
We use   $\mathrm{DWConv}$  to avoid    inter-channel  mixing   at  early stage  where a  $7{\times}7$ convolution   is used.
%
To further reduce the number of learnable weights, we employ zero bias in a set of $C'$   standard $1{\times}1$ convolution filters, denoted  by $\mathrm{Conv}_{1\times1}$, and each filter subsequently requires $C$ weights.
%
%
 CWA provides per-descriptor independence via grouped filtering  ($\mathrm{DWConv}$) and  a simple channel alignment through a $C{\rightarrow}C^\prime$ linear projection ($\mathrm{Conv}_{1\times1} $).

\noindent\textbf{Main Fusion Architecture.} 
We are now ready to explain the fusion module building upon FB and CWA.
Taking as input the  luminance (or chrominance) guidance stack $\mathcal{S} \in \mathbb{R}^{H\times W\times K}$,  it starts from linearly projecting it to obtain $C$ feature maps through 
the standard $1{\times}1$ convolution with bias.
The projected features are then fed into a sequence of FBs, resulting in  $\bar{\mathcal{S}}\in\mathbb{R}^{H\times W\times C}$, expressed as
\begin{equation}
    \bar{\mathcal{S}}  = \mathrm{FB}^T \circ \mathrm{Conv}_{1\times1}(\mathcal{S}),
\end{equation}
where $\mathrm{FB}^T$ denotes the composition of $T$ stacked FB modules, applied  sequentially. 
In parallel, $\mathcal{S}$ is processed by the attention mechanism CWA  to produce an attention map used to mask $\bar{\mathcal{S}} $.
Subsequently, the masked features are  processed by $T$ additional sequential FBs, followed by a final $1{\times}1$ convolution.
This leads to the fusion module below, as
%
\begin{equation}
\label{eq:fusion}
    f(\mathcal{S}) = \mathrm{Conv}_{1\times1} \circ \mathrm{FB}^T \left(\text{CWA}(\mathcal{S})  \odot \bar{\mathcal{{S}}} \right).
\end{equation}
To produce the luminance and reflectance corrections, i.e., $\boldsymbol{\Delta}_L$ and $\boldsymbol{\Delta}_R$, the final $\mathrm{Conv}_{1\times1}$ uses $C=1$ and $C=3$ filters, respectively.
For network training, we employ a hybrid loss, as a weighted sum of a pixel-wise mean squared error (MSE) loss, a multi-scale structural similarity index measure (MS-SSIM) loss \cite{MSSSIM}, and a perceptual loss \cite{PercLoss}, explained  in Section \ref{supp_loss} of supplementary material.

We instantiate this fusion design by Multinex variants.
\emph{Lightweight} uses the full fusion module of Eq.~(18) with $T=3$ FB layers per FB stage.
%
%
\emph{Nano} uses $T=1$ and a simplified fusion path using only the latter FB, which, in turn, retains only its latter MSEF.
All variants share the same enhancement formulation and analytic guidance stacks, while progressively reducing fusion depth, spatial resolution, and module complexity.

\begin{figure*}
    \centering
    \includegraphics[width=1.0\linewidth]{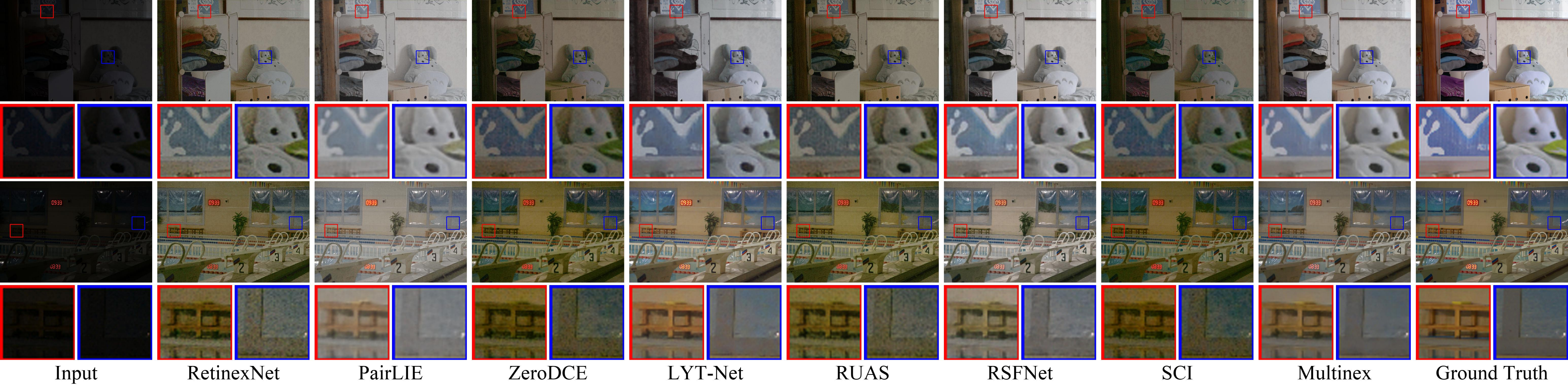}
    \vspace{-18pt}
    \caption{Qualitative comparison of SOTA light-weight approaches~\cite{RetinexNet, pairlie, zero_dce, brateanu2025lyt, RUAS, RSFNet, SCI} and Multinex on reference data.}
    \vspace{-6pt}
    \label{fig:qualitative_LOL}
\end{figure*}

\begin{table*}[t]
\scriptsize
\centering
\renewcommand{\arraystretch}{1.1}
\setlength{\tabcolsep}{3pt}
\resizebox{1.0\textwidth}{!}{
\begin{tabular}{l cc cc cc cc cc cc}
\toprule
& \multicolumn{2}{c}{\textbf{MEF}} 
& \multicolumn{2}{c}{\textbf{LIME}} 
& \multicolumn{2}{c}{\textbf{DICM}} 
& \multicolumn{2}{c}{\textbf{NPE}} 
& \multicolumn{2}{c}{\textbf{Mean}} 
& \multicolumn{2}{c}{\textbf{Model Size}} \\
\cmidrule(lr){2-3}
\cmidrule(lr){4-5}
\cmidrule(lr){6-7}
\cmidrule(lr){8-9}
\cmidrule(lr){10-11}
\cmidrule(lr){12-13}
\multirow{-2.5}{*}{\textbf{Method}}
& \textbf{NIQE}$\downarrow$ & \textbf{BRISQUE}$\downarrow$
& \textbf{NIQE}$\downarrow$ & \textbf{BRISQUE}$\downarrow$
& \textbf{NIQE}$\downarrow$ & \textbf{BRISQUE}$\downarrow$
& \textbf{NIQE}$\downarrow$ & \textbf{BRISQUE}$\downarrow$
& \textbf{NIQE}$\downarrow$ & \textbf{BRISQUE}$\downarrow$
& \textbf{Params (M)}$\downarrow$ & \textbf{GFLOPs}$\downarrow$ \\
\midrule
SNR-Net~\cite{SNR-Net}
& 4.14 & 31.28
& 5.51 & 39.22
& 4.62 & 37.35
& 4.36 & 26.65
& 4.66 & 33.63
& 26.35 & 4.01 \\

LLFlow~\cite{LLFlow}
& 3.92 & 30.27
& 5.29 & 27.06
& {3.78} & 26.36
& 4.16 & 28.86
& 4.29 & 28.14
& 17.42 & 358.40 \\


CIDNet~\cite{yan2025hvi}
& {3.56} & {13.77}
& {4.13} & {16.25}
& 3.79 & {21.47}
& \underline{3.74} & {18.92}
& {3.81} & {17.60}
& 1.88 & 7.57 \\

ZeroDCE~\cite{zero_dce}
& 4.93 & {17.32}
& 5.82 & 20.44
& 4.58 & 27.56
& 4.53 & 20.72
& 4.97 & 21.51
& 0.075 & 4.83 \\

\rowcolor{cvprblue!14}
\textbf{Multinex}
& {3.46} & {13.82}
& \underline{3.76} & {14.60}
& \underline{3.64} & \textbf{16.39}
& \textbf{3.71} & \textbf{12.52}
& \underline{3.64} & \textbf{14.33}
& 0.0446 & 2.50 \\

\rowcolor{cvprblue!14}
\textbf{Multinex$^\dagger$}
& \textbf{3.19} & \textbf{11.89}
& \textbf{3.62} & \textbf{13.39}
& \textbf{3.57} & \underline{17.02}
& {3.82} & \underline{13.09}
& \textbf{3.53} & \underline{15.53}
& 0.0446 & 2.50 \\

\rowcolor{cvprblue!14}
\textbf{Multinex-Nano}
& 3.49 & 15.06
& {3.85} & {15.17}
& 3.90 & {19.97}
& 4.05 & {15.42}
& 3.82 & {16.41}
& 0.0007 & 0.04 \\

\rowcolor{cvprblue!14}
\textbf{Multinex-Nano$^\dagger$}
& \underline{3.45} & \underline{12.55}
& {3.92} & \underline{14.54}
& 3.65 & {17.40}
& 3.94 & {13.67}

& 3.67 & {16.05}
& 0.0007 & 0.04 \\
\bottomrule
\end{tabular}
}
\vspace{-6pt}
\caption{No-reference results on MEF, LIME, DICM, and NPE using NIQE~\cite{NIQE} and BRISQUE~\cite{BRISQUE}. Lower is better. $^\dagger$ indicates results obtained with newly trained models using the same architecture, training configuration, and evaluation protocol as the conference version.}
\vspace{-17pt}
\label{tab:noref_unified}
\end{table*}

\vspace{-3pt}
\section{Experiments and Results}
\vspace{-2pt}



%
We evaluate  Multinex variants, referred to as \emph{Multinex} (44.7K parameters) and \emph{Multinex-Nano} (0.7K parameters), across standard LLIE benchmarks and a downstream task.
%
We use \emph{reference datasets} with paired low/normal-light images, comprising LOL-v1~\cite{RetinexNet} (485:15), LOL-v2~\cite{Sparse} in both real (689:100) and synthetic (900:100) forms, where ($n$:$m$) denotes the train-test split.
To evaluate generalization under uncontrolled lighting, we further test on \emph{no-reference datasets}, including MEF~\cite{MEF}, LIME~\cite{LIME}, DICM~\cite{DICM}, and NPE~\cite{NPE}.
Following common LLIE protocol \cite{Retinexformer,yan2025hvi,GLARE}, we report PSNR, SSIM, and LPIPS on reference datasets, and NIQE~\cite{NIQE} and BRISQUE~\cite{BRISQUE} on no-reference datasets.
In addition to results reported in this section, we include more in Sections \ref{app:ablation} and \ref{app:examples} of supplementary material, showing additional ablation studies,   example demonstration, and challenging cases.

\subsection{Low-Light Image Enhancement}
\label{exp:res}

\noindent\textbf{Quantitative Reference Benchmark.}
We group LLIE models by size into heavy ($>10$M), mid-size ($1$--$10$M), lightweight ($<1$M), and micro ($<10$K).
%
%
Tab.~\ref{tab:llie_vertical_grouped} compares Multinex variants with existing  LLIE methods, where larger models are included to indicate performance gap.
Multinex achieves the strongest results among lightweight models and remains close to the best mid-sized methods at a fraction of their cost.
%
%
Multinex-Nano performs the best within the micro regime while preserving competitive enhancement quality for extreme edge settings.

\noindent\textbf{Quantitative No-reference Benchmark.} 
We evaluate top LLIE models on no-reference benchmarks using NIQE and BRISQUE, and compare the results in
Tab.~\ref{tab:noref_unified}.
Multinex variants achieve the best overall no-reference performance, with mean NIQE 3.66 and BRISQUE 11.91.
Compared with the top-performing CIDNet, it improves mean NIQE by 0.15 and BRISQUE by 5.69 while using $<2.5\%$ parameters.
This shows strong perceptual quality and good generalization to unseen low-light environments.

\noindent\textbf{Qualitative Results.}
%
Fig.~\ref{fig:qualitative_LOL} compares representative outputs from  top lightweight and micro LLIE models.
RetinexNet~\cite{RetinexNet}, ZeroDCE~\cite{zero_dce}, RUAS~\cite{RUAS}, and SCI~\cite{SCI} often leave noise or color artifacts.
PairLIE~\cite{pairlie} and RSFNet~\cite{RSFNet} suppress noise better but tend to weaken color richness.
LYT-Net~\cite{brateanu2025lyt} improves both, yet can still show mild exposure bias.
Multinex better matches the ground truth, e.g., by recovering cleaner illumination gradients and more stable colors.

\subsection{Low-Light Object Detection}

We further evaluate Multinex on low-light object detection using ExDark~\cite{Exdark}.
Table~\ref{tab:exdark_det} reports per-class AP and mAP50 across multiple LLIE models. Here, Multinex is used as pre-encoder for YOLOv3~\cite{yolov3}.
We compare with prior low-light detection and enhancement-based pipelines, including KinD~\cite{KinD}, ZeroDCE~\cite{zero_dce}, PairLIE~\cite{pairlie}, LLFlow, DENet~\cite{denet}, PE-YOLO~\cite{peyolo}, IAT~\cite{iat}, and EMV-YOLO~\cite{emvyolo}.
 Multinex variants show strong performance, giving the best overall results.
%
In particular, Multinex-Nano reaches the highest mAP50   using only 0.7K parameters, showing that the proposed guidance stacks remain effective even at extreme model scale.
\begin{table}[t]
\centering
\renewcommand{\arraystretch}{1.1}
\setlength{\tabcolsep}{1pt}
\resizebox{0.48\textwidth}{!}{%
\tiny
\begin{tabular}{l cccccccccccc c c}
\toprule
\textbf{Method} 
& \textbf{Bic.} & \textbf{Boa.} & \textbf{Bot.} & \textbf{Bus} & \textbf{Car} & \textbf{Cat} & \textbf{Cha.} & \textbf{Cup} & \textbf{Dog} & \textbf{Mot.} & \textbf{Ppl.} & \textbf{Tab.} 
& \textbf{mAP50\%} 
& \textbf{P (K)} \\
\midrule
YOLOv3~\cite{yolov3}
& 79.8 & 75.3 & 78.1 & 92.3 & 83.0 & 68.0 & 69.0 & 79.0 & 78.0 & 77.3 & 81.5 & 55.5
& 76.4 & -- \\

KinD~\cite{KinD}
& 80.9 & 75.0 & 75.8 & 93.3 & 82.4 & 69.4 & 69.2 & 79.0 & 76.9 & 76.3 & 79.6 & 55.4
& 76.1 & 8000 \\

ZDCE~\cite{zero_dce}
& 81.2 & 75.0 & 75.7 & 93.4 & 83.2 & 67.7 & 70.2 & 76.4 & 74.1 & 77.7 & 81.3 & 55.5
& 75.9 & 79 \\

Pair~\cite{pairlie}
& 80.8 & 78.3 & 76.8 & 90.5 & 84.5 & 66.8 & 69.1 & 75.6 & 78.9 & 73.7 & 80.3 & 54.5
& 75.8 & 342 \\


DENet~\cite{denet}
& 80.9 & 79.2 & \underline{80.1} & 90.7 & 84.5 & 70.7 & 72.0 & 79.3 & 80.1 & 76.7 & 82.4 & 58.0
& 77.9 & 45 \\

PE-Y~\cite{peyolo}
& \textbf{84.7} & 79.2 & 79.3 & 92.5 & 83.9 & 71.5 & 71.7 & 79.7 & 79.7 & 77.3 & 81.8 & 55.3
& 78.0 & 91 \\

IAT~\cite{iat}
& 79.8 & 76.9 & 78.6 & 92.5 & 83.8 & 73.6 & 72.4 & 78.6 & 79.0 & 79.0 & 81.1 & 57.7
& 77.8 & 91 \\

EMV-Y~\cite{emvyolo}
& 82.8 & \underline{79.7} & 79.8 & 94.1 & \underline{84.7} & 74.3 & 74.1 & \textbf{83.1} & \textbf{82.7} & 78.1 & 83.6 & \textbf{59.3}
& \underline{79.7} & 27 \\

\rowcolor{cvprblue!14} \textbf{Multinex}
& 81.9 & \textbf{80.1} & 77.9 & \textbf{95.2} & \textbf{86.2} & 73.7 & \underline{75.4} & 80.3 & 82.6 & 79.6 & \underline{84.2} & 58.5
& \underline{79.7} & 45 \\


\rowcolor{cvprblue!14} \textbf{M.-Nano}
& \underline{83.3} & 79.4 & \textbf{81.1} & \underline{94.6} & \underline{84.7} & \textbf{78.3} & \textbf{77.2} & \underline{82.5} & \textbf{82.7} & \textbf{80.7} & \textbf{84.6} & \underline{58.9}
& \textbf{80.7} & 0.7 \\
\bottomrule
\end{tabular}%
}
\vspace{-10pt}
\caption{Low-light object detection on ExDark~\cite{Exdark}: per-class AP\% and overall \textbf{mAP50\%} , with LLIE model parameter count (\textbf{P}). Best results are in \textbf{bold}, second best are \underline{underlined}.}
\label{tab:exdark_det}
\vspace{-15pt}
\end{table}

\begin{table*}[!ht]
\small
\centering
\renewcommand{\arraystretch}{1.0}
\setlength{\tabcolsep}{3pt}
\resizebox{\textwidth}{!}{%
\begin{tabular}{
    lcc
    c  
    cccc
    c  
    lcc
}
\toprule
\multicolumn{3}{c}{\textbf{(a) Multinex Priors}} &
\multicolumn{1}{c}{} & 
\multicolumn{4}{c}{\textbf{(b) CWA/MSEF Analysis}} &
\multicolumn{1}{c}{} & 
\multicolumn{3}{c}{\textbf{(c) CWA and Attention Variants}} \\
\cmidrule(lr){1-3}\cmidrule(lr){5-8}\cmidrule(lr){10-12}
Formulation & H-Dim & PSNR $\uparrow$ &
& 
CWA & MSEF & H-Dim & PSNR $\uparrow$ &
& 
Attention Operation & H-Dim & PSNR $\uparrow$ \\
\midrule
$\hat{\mathbf{I}}_i = f_i(\mathbf{I}_i)$ & 87 & 14.15 & & \xmark & \xmark & 55 & 19.92 & & CBAM$_\text{S}$ & 39 & 20.44 \\
$\hat{\mathbf{I}}_i = f_{\mathcal{L}}(\mathcal{S}_\mathcal{L}, \theta_\mathcal{L}) + \mathbf{I}_i$ & 87 & 20.57 & & \checkmark & \xmark & 55 & 21.34 & &  MHSA + Pooling & 39 & 21.67 \\
$\hat{\mathbf{I}}_i = f_{\mathcal{R}_i}(\mathcal{S}_\mathcal{R}, \theta_\mathcal{R}) + \mathbf{I}_i$  & 87 & 18.50 & & \xmark & \checkmark & 39 & 22.27 & & MDTA & 39 & 22.39 \\
\rowcolor{cvprblue!14} $\hat{\mathbf{I}}_i =f_{\mathcal{L}}(\mathcal{S}_\mathcal{L}, \theta_\mathcal{L}) \odot f_{\mathcal{R}_i}(\mathcal{S}_\mathcal{R}, \theta_\mathcal{R}) + \mathbf{I}_i$ & 39 & 23.19 & & \checkmark & \checkmark & 39 & 23.19 & & CWA & 39 & 23.19 \\
\bottomrule 
\end{tabular}
}
\vspace{-5pt}
\caption{Results of ablation studies (a), (b) and (c) validating design elements of Multinex.}
\vspace{-8pt}
\label{tab:ablation_table_merged}
\end{table*}


\subsection{Ablation Studies}
\label{exp:abl}

To examine the role of different  design elements of Multinex, we perform ablation studies using the  LOL-v1  dataset  under the same size configuration of 45K parameters using the  PSNR metric, and report  results   in Tab.~\ref{tab:ablation_table_merged}.

\begin{figure}
    \centering
    \includegraphics[width=1.0\linewidth,trim={0cm 0cm 0 0cm},clip]{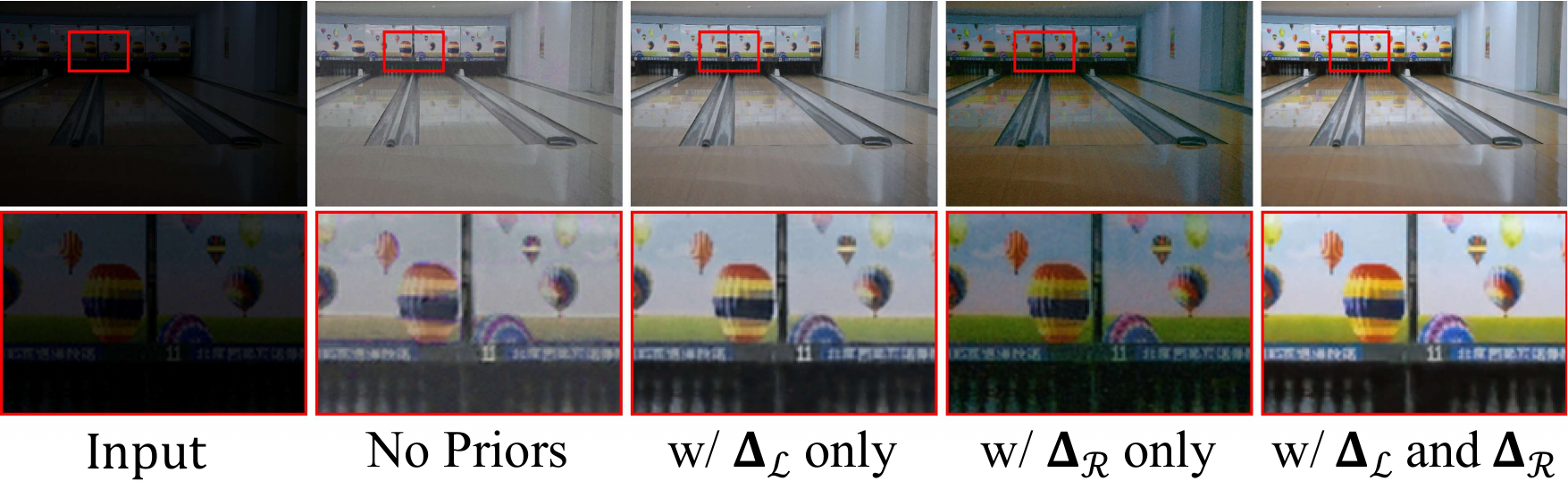}
    \vspace{-18pt}
    \caption{Qualitative ablation on Multinex Priors study (a).}
    \vspace{-15pt}
    \label{fig:ablation_qualitative}
\end{figure}

\noindent\textbf{(a) Multinex Priors.}
We first examine the effect of   Multinex luminance and reflectance. 
Without using both, i.e., $\hat{\mathbf{I}}_i = f(\mathbf{I}_i)$, the network relies solely on raw RGB correlations (no priors) and produces a low 14dB PSNR.  
By using the Multinex luminance (w/ $\boldsymbol\Delta_\mathcal{L}$ only) to adjust, i.e., $\hat{\mathbf{I}}_i = f_{\mathcal{L}}(\mathcal{S}_\mathcal{L}, \theta_\mathcal{L}) + \mathbf{I}_i$,  the PSNR  improves to over 20dB, as explicit brightness cues can guide exposure recovery through illumination-aligned priors.  
By using the Multinex reflectance (w/ $\boldsymbol\Delta_\mathcal{R}$ only) to adjust, i.e., $\hat{\mathbf{I}}_i = f_{\mathcal{R}_i}(\mathcal{S}_\mathcal{R}, \theta_\mathcal{R}) + \mathbf{I}_i$, the PSNR   reaches   18.5dB.
This indicates that color-difference information can stabilize  hue   but is less efficient than illumination structure.  
By combining both Multinex terms (w/ $\boldsymbol\Delta_\mathcal{L}$ and $\boldsymbol\Delta_\mathcal{R}$) as in our Eq. (\ref{eq:final_fusion_stack}),  the PSNR further improves to over 23dB, forming a physically consistent decomposition.
This shows that Eq. (\ref{eq:final_fusion_stack}) enables Multinex   to learn suitable illumination and color cues that correctly adjust to the true luminance and reflectance.
Fig. \ref{fig:ablation_qualitative} demonstrates examples resulted from these settings.
%

\noindent\textbf{(b) CWA/MSEF Analysis.}
We propose CWA and FB supported by its core operation MSEF to construct  the fusion module in Eq. (\ref{eq:fusion}).  
We thus assess the effect of CWA and MSEF by removing these   from the module.  
Removing both results in a PSNR reduction from  23dB to  20dB, despite greater depth within each FB layer.
Using CWA alone improves to 21dB, while   MSEF alone improves to  22dB, benefiting from relevance weighting of analytic descriptors for CWA, and from local detail enhancement by excitation and depthwise spatial filtering for MSEF.
Further performance improvement enabled by using both CWA and MSEF confirms that MSEF (localized features) and CWA (channel re-weighting) can complement each other effectively.


\noindent\textbf{(c) CWA and Attention Variants.}
To examine the effectiveness of our attention design, we compare CWA with alternative  attention formulations by replacing.
These include the convolutional block attention module, specifically the spatial branch CBAM$_\text{S}$~\cite{CBAM},  multi-head self-attention (MHSA) with pooling~\cite{brateanu2025lyt, SNR-Net},  and   multi-Dconv head transposed attention (MDTA) that improves the vanilla MHSA~\cite{Restormer}.  
%
%
CBAM$_\text{S}$ reaches a PSNR of  20dB, which  entangles per-component responses, weakening component specificity.  
MHSA with pooling improves to just under 22 dB, but it loses fine structure due to pooling before attention. 
MDTA   surpasses 22 dB, confirming the utility of transposed self-attention, but still mixes component channels.  
CWA improves component selectivity at a low complexity,  producing independent, per-component  attention maps for each descriptor in a stack, achieving the best PSNR of 23dB.



\vspace{-3pt}
\section{Conclusion, Limitation and Future Work }
\vspace{-2pt}
\label{sec:conclusion}

We have proposed Multinex, a compact and physically grounded LLIE framework that introduces an analytic luminance-reflectance delta decomposition for image correction, and further strengthens this mechanism via lightweight, learnable fusion modules.
By reformulating the image enhancement as an additive residual process rather than a full reconstruction task, Multinex achieves strong illumination and color fidelity while requiring only a fraction of the parameters used by existing approaches.
Comprehensive experiments on  (un)paired   benchmarks confirm that Multinex is able to deliver  SOTA performance at real-time cost, demonstrating the  power of combining analytic priors and modern lightweight design, offering on-edge practicality and physical explainability.

The Multinex architecture has largely reduced parameter consumption while maintaining  enhancement quality.
However, like  other LLIE techniques,  it can still struggle with complex spectral distortions, lens flares, or mixed artificial-natural lighting. 
For extremely dark scenes, the analytic representation priors may amplify sensor noise.
The  additive residual is highly stable for under-exposure, but   can become less  effective with clipped highlights or high dynamic range (HDR) recovery.  
Future work will extend Multinex to  these particularly  challenging cases,  by leveraging the potential of alternative frameworks, such as transformers or probabilistic diffusion, by applying Multinex principles to other domains such as intrinsic image decomposition, color constancy, and underwater or haze removal, and by exploring tone-mapping and multiplicative formulations as alternatives to the additive residual.

\section*{Acknowledgements}
Part of this research was supported by the project "Romanian Hub for Artificial Intelligence - HRIA", Smart Growth, Digitization and Financial Instruments Program, 2021-2027, MySMIS, no.351416.

{
    \small
    \bibliographystyle{ieeenat_fullname}
    \bibliography{ref}
}

\clearpage
\setcounter{page}{1}
\maketitlesupplementary


\def\thesection{\Alph{section}}
\setcounter{section}{0}

\textbf{Content Summary:}  Section \ref{sec:supp_lra} explains the linear reconstruction analysis (LRA), which  is used to  visualize feature maps.
Secion \ref{sec:supp_guidance_study} includes further studies on additional analytic representation priors used by the luminance and reflectance guidance stacks, and explains the reason for choosing the used priors by Multinex.
Section \ref{app:multinex_detail} provides additional details regarding to  Multinex to complement the main paper, including its basic building blocks, the hybrid loss function, as well as the  implementation and evaluation considerations.
Section  \ref{app:ablation} presents extra experimental results and discussions on additional ablation studies of the adopted losses and model components, while  Section \ref{app:examples} presents expanded qualitative visualizations.

%
%


\section{ Linear Reconstruction Analysis}
\label{sec:supp_lra}

Given an input feature map $\mathbf{X} \in \mathbb{R}^{H\times W \times K}$ and a target signal $\mathbf{T}\in \mathbb{R}^{H\times W \times C}$ defined according to the analysis goal, LRA examines how much information in $\mathbf{T}$ is retained by $\mathbf{X}$.
%
The core idea is to assess \emph{what portion of the target signal can be reconstructed through a linear combination of the principal components of the feature map}.
%

%

%
 
\paragraph{Pre-processing.} We  pre-process the data before performing LRA. 
We firstly flatten the feature map into $\mathbf{X}_f \in \mathbb{R}^{N \times K}$, where $N = HW$. 
Each row of $\mathbf{X}_f$ corresponds to a pixel   and each column to a feature type (channel).
We then center each column by subtracting its mean, obtaining $\mathbf{X}_c = \mathbf{X}_f - \mu_\mathbf{X}$ with $\mu_\mathbf{X} = \tfrac{1}{N}\mathbf{1}^\top \mathbf{X}_f$, where $\mathbf{1}$ is a length-$N$ column vector with all elements equal to 1.
%
Similarly, we flatten the target signal into $\mathbf{T}_f \in \mathbb{R}^{N \times C}$, where $C$ depends on the target definition (e.g., $C{=}3$ for RGB and $C{=}1$ for luminance), and center its columns by subtracting the mean vector $\mu_\mathbf{T} = \tfrac{1}{N}\mathbf{1}^\top \mathbf{T}_f$ to obtain $\mathbf{T}_c = \mathbf{T}_f - \mu_\mathbf{T}$.

\paragraph{LRA.} It first applies principal component analysis (PCA), linearly projecting $\mathbf{X}_c$ onto its top $D$ principal components through the orthogonal projection matrix $\mathbf{P}_{\mathrm{PCA}} \in \mathbb{R}^{K \times D}$.
This yields the reduced feature matrix $\mathbf{Z} \in \mathbb{R}^{N \times D}$,
\begin{equation}
  \mathbf{Z} = \mathbf{X}_c \mathbf{P}_{\mathrm{PCA}},
\end{equation}
which captures the dominant variation across the feature channels.
PCA  acts as a compact linear representation that removes redundancy in $\mathbf{X}$ while preserving its major structure.
We then fit a linear model to map $\mathbf{Z}$ to the centered target $\mathbf{T}_c$ through ridge regression, resulting in the  following mapped target: 
\begin{equation}
  \hat{\mathbf{T}}_c = \mathbf{Z}\mathbf{W},
\end{equation}
where $\mathbf{W} \in \mathbb{R}^{D \times C}$ is computed by
\begin{equation}
  \mathbf{W}
  = (\mathbf{Z}^\top \mathbf{Z} + \lambda \mathbf{I}_D)^{-1}\mathbf{Z}^\top \mathbf{T}_c,
\end{equation}
with $\mathbf{I}_D$ being the identity matrix of size $D$ and $\lambda > 0$ a small ridge regularization parameter.
Using $\textmd{reshape}(\cdot)$ to restore  a flattened signal to spatial dimensions $H \times W \times C$, the  final reconstructed target is given by
\begin{equation}
  \hat{\mathbf{T}}(\mathbf{X}) = \textmd{reshape}(\mathbf{Z}\mathbf{W} + \mu_{\mathbf{T}}).
\end{equation}
 
\paragraph{Feature Visualization.}  By setting  $\mathbf{T}$  as the low-light image input, how different the reconstructed image $\hat{\mathbf{T}}(\mathbf{X})$ is from  $\mathbf{T}$  provides a simple and interpretable measure of the information content carried by the feature map.
Therefore, visualizing the reconstructed image $\hat{\mathbf{T}}(\mathbf{X})$   helps establish an understanding of the physical role of the feature map $\mathbf{X}$ for enhancing $\mathbf{T}$.



\section{More Studies On Representation Prior}
\label{sec:supp_guidance_study}

\begin{figure}
    \centering
    \includegraphics[width=0.5\linewidth]{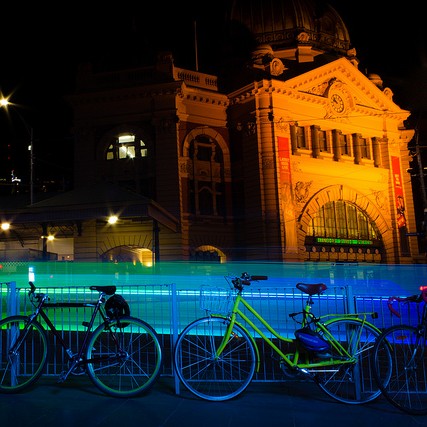}
    \caption{The example image  used by Sec. \ref{sec:supp_guidance_study}.}
    \vspace{-10pt}
    \label{fig:suppl_input}
\end{figure}

\begin{figure*}
    \centering
    \includegraphics[width=1.0\linewidth]{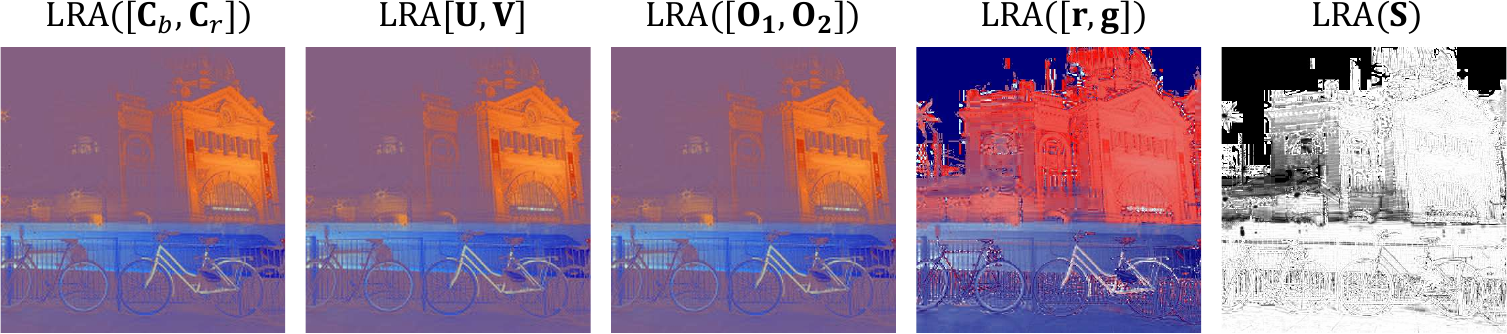}
    \caption{LRA visualization of chrominance candidates  $\{[\mathbf{C}_b,\mathbf{C}_r], [\mathbf{U}, \mathbf{V}], [\mathbf{O}_1, \mathbf{O}_2], [\mathbf{r}, \mathbf{g}], \mathbf{S}\}$.}
    \vspace{-10pt}
    \label{fig:supp_pseudo_rgb_ridge}
\end{figure*}

\begin{figure}
    \centering
        \includegraphics[width=1.0\linewidth]{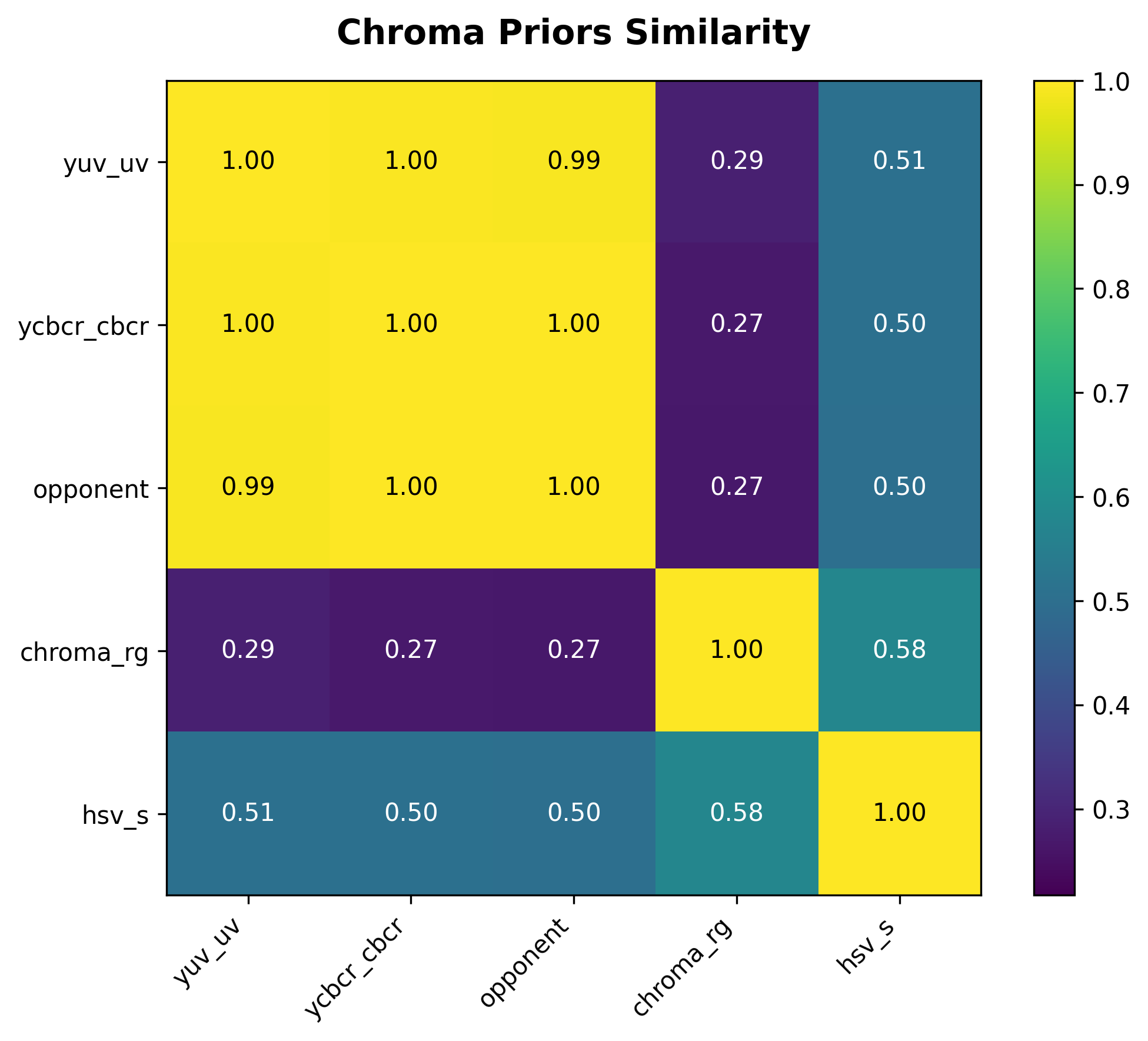}
    \caption{Similarity heatmaps between all candidate grouped reflectance-guidance components.}
    \label{fig:supp_chroma_corr}
\end{figure}

\begin{figure*}
    \centering
    \includegraphics[width=1.0\linewidth]{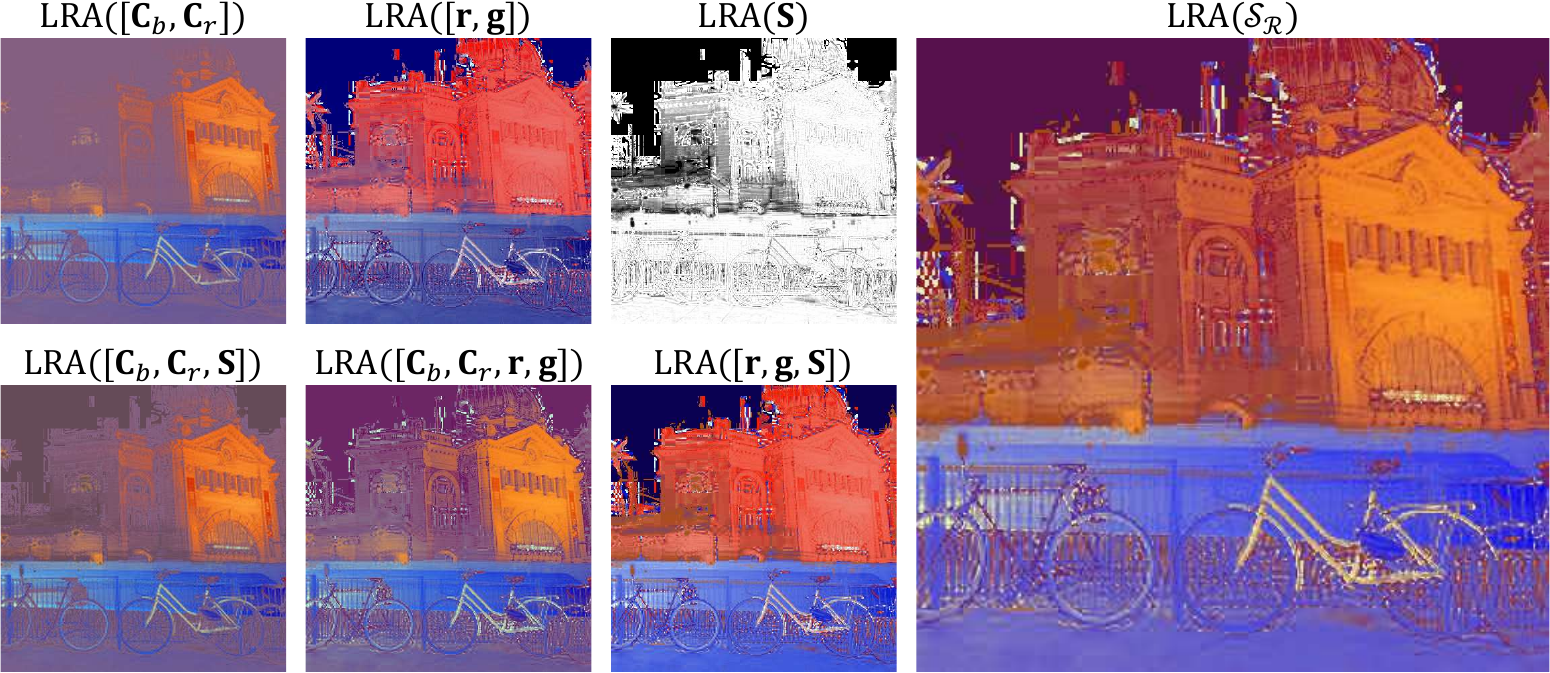}
    \caption{LRA visualization of combinations of the chosen feature maps  for the  reflectance guidance stack $\mathcal{S}_\mathcal{R}$.}
    \vspace{-10pt}
    \label{fig:lra_all}
\end{figure*}


We have proposed the luminance and reflectance guidance stacks in Eqs.~(\ref{eq:illum_stack}) and  ~(\ref{eq:reflectance_stack}), serving as the representation prior.
To arrive at this design,  we initially considered a larger pool of analytic luminance and chrominance descriptors for computing the feature maps, including multiple linear and nonlinear ones.
Strong non-linearity typically has the potential to encode richer or more distinctive relationships.
Therefore, our   selection strategy  prioritizes features that are  (1) inherently non-linear, and/or (2) less correlated  with the other features.
The goal is to enable the finally selected guidance stacks to offer the widest possible range of complementary  lenses for the model to view the image.
For each guidance stack, we explain below the additional feature candidates considered during model design.
We then analyze and visualize the  contribution of both the selected and non-selected descriptors by LRA, to validate our representation-prior design.
The image shown in Fig.~\ref{fig:suppl_input} is used throughout the analysis.
%
%
%


\subsection{Reflectance Guidance Stack}
\label{sec:supp_reflectance_study}

\subsubsection{Additional Chrominance Candidates }
\label{sec:supp_extended_candidates_C}

For the reflectance guidance stack, besides $\{\mathbf{C}_b,\mathbf{C}_r,\mathbf{r},\mathbf{g},\mathbf{S}\}$ in Eq.~(\ref{eq:reflectance_stack}), we considered two further pairs of chroma descriptors. These include the YUV chroma pair $[\mathbf{U},\mathbf{V}]$, computed as
\begin{align}
  \mathbf{U} &= -0.14713 \mathbf{I}_R - 0.28886 \mathbf{I}_G + 0.43600 \mathbf{I}_B, \\
  \mathbf{V} &= \phantom{-}0.61500 \mathbf{I}_R - 0.51499 \mathbf{I}_G - 0.10001 \mathbf{I}_B,
\end{align}
and the pair of the first two opponent channels $[\mathbf{O}_1,\mathbf{O}_2]$,  as 
\begin{align}
  \mathbf{O}_1 &= \frac{1}{\sqrt{2}}(\mathbf{I}_R - \mathbf{I}_G), \\
  \mathbf{O}_2 &= \frac{  1}{\sqrt{6}} (\mathbf{I}_R + \mathbf{I}_G - 2 \mathbf{I}_B).
\end{align}
The pair $[\mathbf{U},\mathbf{V}]$ encodes blue-yellow and red-cyan differences in a luminance-decoupled fashion, while $[\mathbf{O}_1,\mathbf{O}_2]$ is derived from the classical opponent color theory, spanning red-green and blue-yellow axes in a normalized manner.
The above, together with Eq.~(\ref{eq:reflectance_stack}), forms an initial candidate pool of chroma descriptors, i.e.,
\begin{equation}
\hat{\mathcal{S}}_\mathcal{C}(\mathbf{I})  = \{\mathbf{C}_b, \mathbf{C}_r, \mathbf{U}, \mathbf{V}, \mathbf{O}_1, \mathbf{O}_2, \mathbf{r}, \mathbf{g}, \mathbf{S}\}.
\end{equation}
%

\subsubsection{Comparative Visual Analysis}
\label{sec:supp_chrom_study_comp}

The three groups of chroma descriptors $[\mathbf{C}_b,\mathbf{C}_r]$,   $[\mathbf{U},\mathbf{V}]$  and $[\mathbf{O}_1,\mathbf{O}_2]$ are linear   combinations of the underlying RGB channels, thus linearly correlated to each other.
So we include only one group to the final chrominance stack.
Below we visually validate our preference of using $[\mathbf{C}_b,\mathbf{C}_r]$ over $[\mathbf{U},\mathbf{V}]$  and $[\mathbf{O}_1,\mathbf{O}_2]$, to complement the   nonlinear descriptors $[\mathbf{r}, \mathbf{g}]$ and $\mathbf{S}$.

We produce the   LRA visualization results  with $D=3$.
Fig.~\ref{fig:supp_pseudo_rgb_ridge} visualizes how well the chrominance candidates  $[\mathbf{C}_b,\mathbf{C}_r]$, $[\mathbf{U}, \mathbf{V}]$, $[\mathbf{O}_1, \mathbf{O}_2]$, $ [\mathbf{r}, \mathbf{g}]$, and $ \mathbf{S}$ can reconstruct the  RGB content of the low-light image $ \mathbf{I}$.
%
To quantify the redundancy among the candidate chroma priors, we compute their pairwise Pearson correlation. 
Because the descriptors vary in channel depth, we first compute the pixel-wise $L_2$ norm across the channel dimension for each prior. 
This reduces every candidate to a single spatial magnitude map representing its overall activation. 
We then flatten these magnitude maps and compute the Pearson correlation coefficient between them, yielding a unified similarity heatmap.
%
%

Fig.~\ref{fig:supp_pseudo_rgb_ridge}  shows that the three linear groups of $[\mathbf{C}_b,\mathbf{C}_r]$, $[\mathbf{U},\mathbf{V}]$, and $[\mathbf{O}_1,\mathbf{O}_2]$  are able to achieve reconstructions with near-identical structure and color fidelity, which is   consistent with their analytic formulations.
The group  analysis in Fig.~\ref{fig:supp_chroma_corr} (\textbf{left}) confirms that $[\mathbf{C}_b,\mathbf{C}_r]$, $[\mathbf{U},\mathbf{V}]$, and $[\mathbf{O}_1,\mathbf{O}_2]$ are highly correlated among themselves.
However, the single-component heatmap in Fig.~\ref{fig:supp_chroma_corr}  (\textbf{right}) reveals that   $[\mathbf{U},\mathbf{V}]$ and $[\mathbf{C}_b,\mathbf{C}_r]$   are   less correlated with the non-linear descriptors $\mathbf{r},\mathbf{g},\mathbf{S}$ as compared to $[\mathbf{O}_1,\mathbf{O}_2]$.
This suggests that $[\mathbf{U},\mathbf{V}]$ and $[\mathbf{C}_b,\mathbf{C}_r]$ provide more complementary information when being combined with the normalized chromaticities and saturation.
%
Between $[\mathbf{U},\mathbf{V}]$ and $[\mathbf{C}_b,\mathbf{C}_r]$, we empirically observe that $[\mathbf{C}_b,\mathbf{C}_r]$ yields slightly better discrimination of yellow-orange hues in the LRA reconstruction, which are particularly relevant to low-light scenes (e.g., street lights and indoor tungsten illumination).
Based on these, for the reflectance guidance stack we retain $\mathbf{C}_b$ and $\mathbf{C}_r$, together with the non-linear maps $\mathbf{r}, \mathbf{g}$ and $\mathbf{S}$, leading to the final design in Eq.~(\ref{eq:reflectance_stack}).

To visualize the effectiveness of $\mathcal{S}_\mathcal{R}$ as guidance for reflectance modeling, we use LRA on prior combinations of the proposed stack. As Fig.~\ref{fig:lra_all} shows, that $[\mathbf{C}_b,\mathbf{C}_r]$ successfully recovers general color of the input, $[\mathbf{r},\mathbf{g}]$ finds regions that are uniform in reflectance, and $\mathbf{S}$ provides greater structural boundary and color control in dark areas.


\subsection{Luminance Guidance Stack}
\label{sec:supp_illum_study}

\subsubsection{Additional Luminance Candidates}
\label{sec:supp_extended_candidates_I}

Besides $\left\{ \mathbf{Y}_\text{Rec.709}, \mathbf{Y}_\text{vmax}, \mathbf{Y}_\text{lightness}, \mathbf{Y}_{\text{L}_2}\right\}$ in Eq.~(\ref{eq:illum_stack}), we also considered the following luminance candidates:
\begin{align}
  &\mathbf{Y}_\text{mean} = \tfrac{1}{3}\left(\mathbf{I}_R + \mathbf{I}_G + \mathbf{I}_B\right), \\
  &\mathbf{Y}_\text{YCgCo} = 0.25 \mathbf{I}_R + 0.50 \mathbf{I}_G + 0.25 \mathbf{I}_B.
\end{align}
Here, $\mathbf{Y}_\text{mean}$ is the simple arithmetic mean intensity, while $\mathbf{Y}_\text{YCgCo}$ is the luminance component of the YCgCo color space, which emphasizes more the green channel   through a physically motivated transform.
Together with Eq.~(\ref{eq:illum_stack}), they form  our initial candidate pool of six luminance  descriptors:
\begin{equation}
\hat{\mathcal{S}}_\mathcal{L}(\mathbf{I})  = \{\mathbf{Y}_\text{Rec.709}, \mathbf{Y}_\text{mean}, \mathbf{Y}_\text{YCgCo},
  \mathbf{Y}_{\text{vmax}}, \mathbf{Y}_{\text{lightness}}, \mathbf{Y}_{\text{L}_2}\}.
\end{equation}

\begin{figure*}
    \centering
    \includegraphics[width=1.0\linewidth]{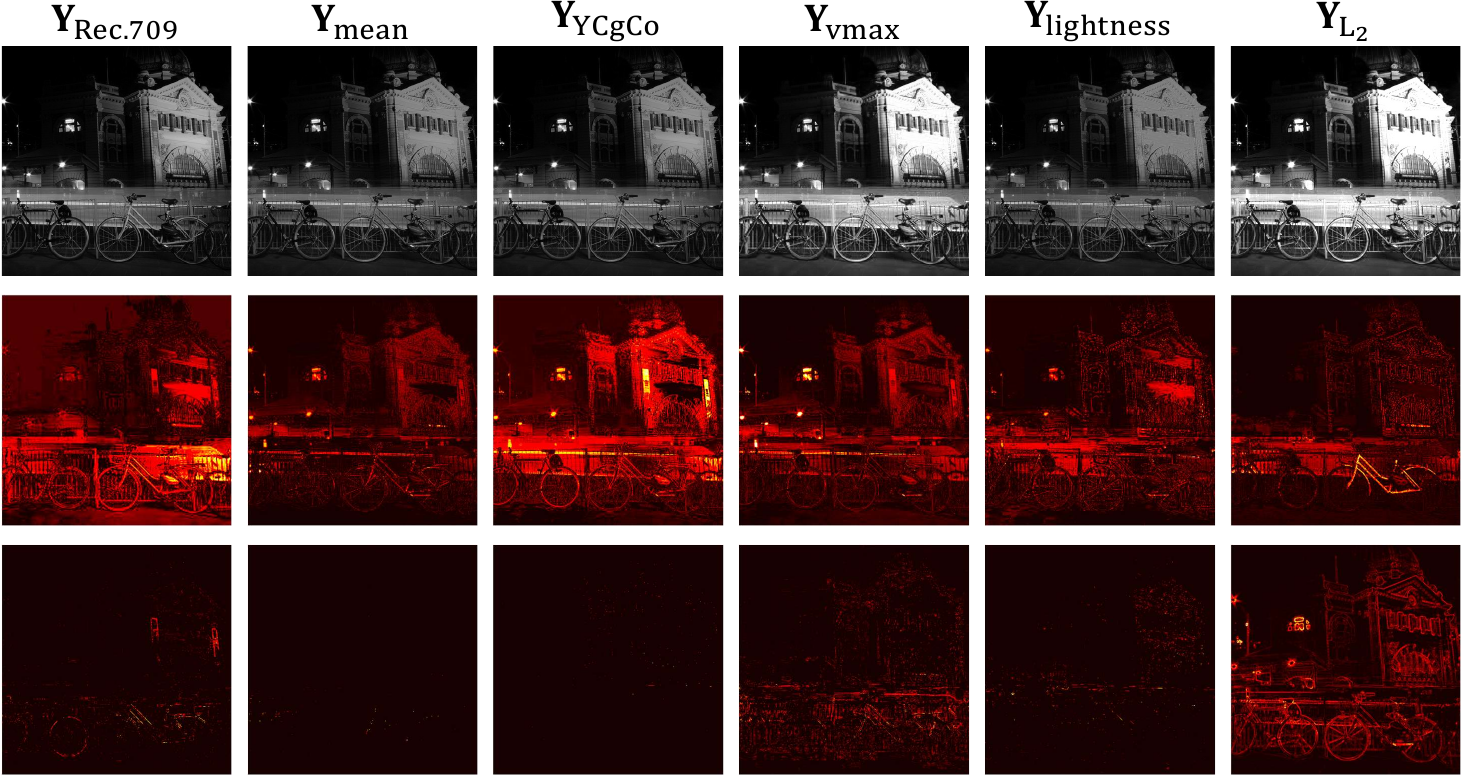}
    \caption{\textbf{Top Row:} Candidate illumination  feature descriptors  $\{\mathbf{Y}_\text{Rec.709}, \mathbf{Y}_\text{mean}, \mathbf{Y}_\text{YCgCo},
  \mathbf{Y}_{\text{vmax}}, \mathbf{Y}_{\text{lightness}}, \mathbf{Y}_{\text{L}_2}\}$. \textbf{Middle Row:} $\boldsymbol{\Delta}_E(c)$ maps of the same candidate descriptors, where $c\in \{1,2,\ldots, 6 \}$. \textbf{Bottom row:} $\boldsymbol{\Delta}_G(c)$ maps of the same candidate descriptors.}
  \vspace{-10pt}
    \label{fig:supp_luma_gray}
\end{figure*}

\begin{figure}
    \centering
    \includegraphics[width=1.0\linewidth]{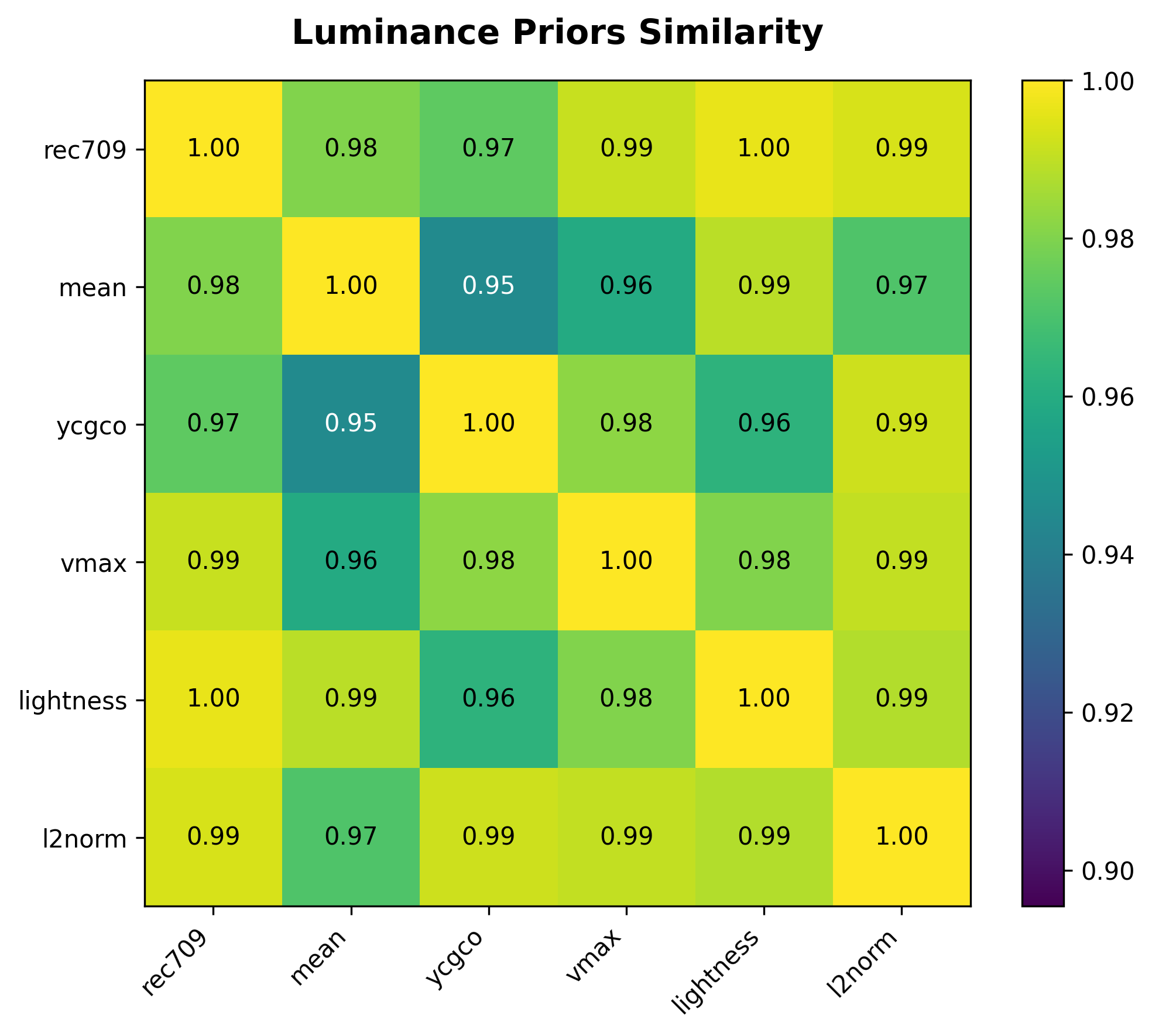}
    \caption{Similarity heatmap between all candidate individual illumination-guidance components.}
    \vspace{-10pt}
    \label{fig:supp_luma_corr}
\end{figure}

\subsubsection{Descriptor Importance Analysis}
\label{supp_illum_importance}

To examine the importance of a luminance descriptor in capturing essential information, we introduce a  leave-one-out approach that measures the loss of expressivity upon removing each descriptor from the full six-prior stack.
As a result, a greater loss indicates a more important  descriptor.
In particular, we measure the loss  across two orthogonal metrics computed pixel wise.

First, we apply a simple edge detection method of maximum gradient operator to highlight the direction along which  the intensity changes the most for each pixel, e.g., by using a sober filter.
Denote the output gradient map for each luminance feature map by $\mathbf{G}_c \in \mathbb{R}^{H\times W}$ for $c\in \{1,2,\ldots, K_\mathcal{L} \}$.
We focus on  the strongest geometric boundary across all the prior candidates, which is computed by applying the max operator element-wise over the gradient maps, as 
%
\begin{equation}
\mathbf{G}_{K_\mathcal{L}}  = \max_{c=1}^{K_\mathcal{L}} \mathbf{G}_c.
\end{equation}
The resulting gradient map extracts structural information from the stack.

Second, we  compute the orthogonal energy for each pixel, which is characterized by a $K_\mathcal{L} $-dimensional vector corresponding to the $K_\mathcal{L}$ candidate maps, measuring the global shading expressivity.
We  apply PCA to  the set of $N=HW$ pixel vectors, and compute the energy using the  non-principal components.
Denote a centered pixel vector by $\mathbf{S}_{ij} \in \mathbb{R}^{K_\mathcal{L}}$, the resulting energy map of all pixels by $\mathbf{E}_{K_\mathcal{L}} \in \mathbb{R}^{H\times W}$, and the $k$-th principal  direction by $\mathbf{p}_k \in \mathbb{R}^{K_\mathcal{L}}$.
Each element of the energy map $\mathbf{E}_{K_\mathcal{L}} $ is then computed by 
\begin{equation}
e_{ij} =\sqrt{\sum_{k=2}^{K_\mathcal{L}} \left(\mathbf{S}_{ij}^T\mathbf{p}_k\right)^2},
\end{equation}
summing over the non-dominant principal directions.



To quantify the unique contribution of each candidate, we   remove each prior from the full stack $\mathcal{S}_\mathcal{L}$ and measure the resulting reduction in the gradient and energy maps.
This process is expressed as below: 
\begin{align}
\boldsymbol{\Delta}_G(c) &=   \max\left(0, \mathbf{E}_{K_\mathcal{L}}  - \mathbf{E}_{K_\mathcal{L} \setminus \{c\}} \right), \\
\boldsymbol{\Delta}_E(c) &=   \max\left(0, \mathbf{G}_{K_\mathcal{L}}  - \mathbf{E}_{K_\mathcal{L} \setminus \{c\}}\right).
\end{align}
The resulting  maps are referred to as the   energy   and gradient importance maps, respectively, and are used for  visualization and analysis.

\subsubsection{Comparative Visual Analysis}
\label{sec:supp_illum_study_comp}

The non-linear luminance descriptors of $\mathbf{Y}_{\text{vmax}}$, $\mathbf{Y}_{\text{lightness}}$, and $\mathbf{Y}_{\text{L}_2}$ provide distinct physical and perceptual interpretations, such as maximum channel response, HSL lightness, and RGB energy, to complement the linear descriptor $\mathbf{Y}_\text{Rec.709}$.
We quantitatively and visually validate our selection of the final four-component stack below, specifically justifying the choice of $\mathbf{Y}_\text{Rec.709}$ over the alternative linear descriptors $\mathbf{Y}_\text{mean}$ and $\mathbf{Y}_\text{YCgCo}$.

The top row of Fig.~\ref{fig:supp_luma_gray} visualizes each candidate illumination map as it is individually. 
Furthermore, we compute the correlation matrix over the six maps, visualized in Fig.~\ref{fig:supp_luma_corr}. As expected, the three linear descriptors of $\mathbf{Y}_\text{Rec.709}$, $\mathbf{Y}_\text{mean}$, $\mathbf{Y}_\text{YCgCo}$ are highly correlated with one another. 
Since they represent correlated transformations of the linear RGB space, we only adopt  one linear descriptor   for the guidance stack.

\begin{table}[h]
\centering
\small.        
\caption{Importance ranking of candidate luminance priors. Higher values indicate greater unique contribution to the stack.}
\label{tab:illum_ranking}
\begin{tabular}{l c c c}
\toprule
Prior & $\boldsymbol{\Delta}_E(c) \uparrow$ / Rank & $\boldsymbol{\Delta}_G(c)\uparrow$ / Rank & Avg. Rank \\
\midrule
$\mathbf{Y}_\text{Rec.709}$ & \textbf{0.0132} / 1 & 0.0007 / 3 & {2.0} \\
$\mathbf{Y}_\text{vmax}$ & 0.0107 / 2 & 0.0019 / 2 & {2.0} \\
$\mathbf{Y}_{\text{L}_2}$ & 0.0012 / 6 & \textbf{0.0194} / 1 & 3.5 \\
$\mathbf{Y}_\text{lightness}$ & 0.0029 / 4 & 0.0002 / 4 & 4.0 \\
$\mathbf{Y}_\text{YCgCo}$ & 0.0038 / 3 & 0.0000 / 6 & 4.5 \\
$\mathbf{Y}_\text{mean}$ & 0.0017 / 5 & 0.0000 / 5 & 5.0 \\
\bottomrule
\end{tabular}
\end{table}

The middle and bottom rows of Fig.~\ref{fig:supp_luma_gray} visualize energy and gradient importance maps for each candidate descriptor.
We also compute scalar energy and gradient  importance scores each computed by averaging the corresponding importance maps over all pixels for each candidate descriptor. 
The two resulting scores are denoted by $\Delta E_c $ and $\Delta G_c$.
We report the scores and the descriptor rankings in Tab.~\ref{tab:illum_ranking}, where higher scores indicate greater expressivity loss, thus higher importance.
The results demonstrate distinct roles among the candidate descriptors, justifying  the necessity of a multi-prior stack. 
Specifically, $\mathbf{Y}_{\text{L}_2}$ dominates in gradient importance, proving critical for preserving high-frequency structural edges.
But it contributes the least to global shading variance. 
Conversely, $\mathbf{Y}_\text{vmax}$ captures the most global illumination variance, excelling at distinguishing bright specularities from diffuse regions, but provides weaker structural guidance. 
Among the highly correlated linear candidates, $\mathbf{Y}_\text{Rec.709}$ significantly outperforms both $\mathbf{Y}_\text{mean}$ and $\mathbf{Y}_\text{YCgCo}$ across both metrics. 
Consequently, we retain $\mathbf{Y}_\text{Rec.709}$ as our sole linear luminance and discard the others, finalizing our four-dimensional illumination guidance stack as in Eq.~\ref{eq:illum_stack}.

\section{Additional Details on Multinex}
\label{app:multinex_detail}

\subsection{Basic Neural Building Blocks}

\label{sec:blocks}
We use  a few basic neural building blocks in our lightweight fusion module.
Taking as input a tensor  $\mathbf{X} \in\mathbb{R}^{H\times W\times C}$, these include operations such as: depth-wise convolution $\text{DWConv}: \mathbb{R}^{H\times W\times C} \rightarrow \mathbb{R}^{H \times W \times C}$ and depth-wise separable convolution $\text{DSConv}: \mathbb{R}^{H\times W\times C} \rightarrow \mathbb{R}^{H \times W\times C}$.
%
%
Both convolutions are designed to reduce the number of filter weights (i.e. parameters) to learn, which fits our goal of constructing a lightweight network. 
Additionally,   two typical activation functions are employed, including the sigmoid activation $\sigma: \mathbb{R}^{H\times W\times C} \rightarrow  [0,1]^{H\times W\times C}$ and ReLU activation $\sigma_{\text{ReLU}}: \mathbb{R}^{H\times W\times C} \rightarrow [0, +\infty)^{H\times W\times C}$.

Another used   building block is the recent  multi-stage squeeze \& excite fusion (MSEF) module,  i.e., a lightweight architecture  with its effectiveness   demonstrated particularly for LLIE
\cite{brateanu2025lyt}, denoted by $\text{MSEF}: \mathbb{R}^{H\times W\times C} \rightarrow \mathbb{R}^{H  \times W \times C }$. 
It generalizes the squeeze-and-excitation (SE) mechanism \cite{hu2018squeeze}.
Given the input feature $\mathbf{X} \in\mathbb{R}^{H\times W\times C}$, it computes an output feature   of the same size $\mathbf{Y} \in\mathbb{R}^{H\times W\times C}$: 
 \begin{equation}
\label{MSEF1}
\mathbf{Y}  = \mathbf{X} +  \text{DWConv} \circ \text{LN}(\mathbf{X})  \odot \mathbf{Z}. 
 \end{equation}
 where $f\circ g(x) = f(g(x))$ denotes the composition of   two functions. 
In order to obtain $\mathbf{Z}\in\mathbb{R}^{H\times W\times C}$,   a set of adaptive channel weights $ \mathbf{w} =[w_1, w_2, \ldots, w_C] \in  \mathbb{R}^C$ are   computed  by a two-layer excitation bottleneck, i.e.,
\begin{equation}
\label{MSEF2}
    \mathbf{w}  = \sigma_{\text{tanh}}\big(\mathbf{W}_2 \sigma_{\text{ReLU}}(\mathbf{W}_1 \text{GAP}\circ\text{LN}(\mathbf{X}))\big),
\end{equation}
where the linear projection matrices $\mathbf{W}_1 \in \mathbb{R}^{d\times C}$ and $\mathbf{W}_2\in \mathbb{R}^{C\times d}$  ($d< C$)  form a feature compression-expansion pair.
Each adaptive weight $w_i$ is then  used to recalibrate the normalized features from the corresponding channel, denoted as $\text{LN}_i(\mathbf{X})$,  by multiplication, i.e.,
\begin{equation}
    \mathbf{Z}_i  = w_i\text{LN}_i(\mathbf{X}), \text{ for }i=1,2,\ldots, C.
\end{equation}
Such a design allows the MSEF module to capture both local fine texture (e.g., through convolution) and global semantics (e.g., through global pooling) in the input, with negligible computational cost.  

\subsection{Multinex Loss Function}
\label{supp_loss}

To train the illumination and reflectance networks $f_{\mathcal{L}}$ and $f_{\mathcal{R}}$ derived from the fusion module, we adopt a hybrid loss that balances pixel-level fidelity, structural consistency, and perceptual quality, expressed as
\begin{equation}
\label{eq:loss_hybrid}
\mathcal{L}
= \lambda_{\text{MSE}} \mathcal{L}_{\text{MSE}}
+ \lambda_{\text{MS-SSIM}} \mathcal{L}_{\text{MS-SSIM}}
+ \lambda_{\text{Perc}} \mathcal{L}_{\text{Perc}}.
\end{equation}
We adopted the existing  hyper-parameter setting \cite{brateanu2025kant} of $\lambda_{\text{MSE}}=1.0$, $\lambda_{\text{MS-SSIM}}=0.2$, and $\lambda_{\text{Perc}}=0.01$ in our implementation,    weighting the contribution of each loss component.
For the sake of convenience, we explain each individual loss term computed over one image pair $(\mathbf{I}_{\text{GT}}, \hat{\mathbf{I}})$, containing the predicted enhanced image $\hat{\mathbf{I}}$ and its corresponding ground-truth well-lit image $\mathbf{I}_{\text{GT}}$.

\begin{figure}
    \centering
    \includegraphics[width=0.8\linewidth]{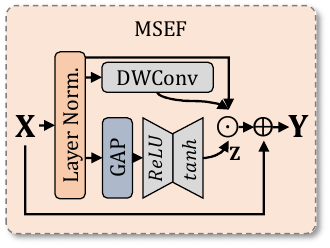}
    \caption{ MSEF module architecture.}
    \label{fig:placeholder}
\end{figure}

\paragraph{MSE Loss.}
The pixel-wise mean squared error (MSE) encourages numerically accurate reconstruction of the enhanced image, defined as 
\begin{equation}
\label{eq:loss_mse}
\mathcal{L}_{\text{MSE}}
= \frac{1}{N} 
\big\|  \hat{\mathbf{I}} - \mathbf{I}_{\text{GT}}  \big\|_2^2,
\end{equation}
where $N$ is the total number of pixels.
It stabilizes training and provides strong gradients for correcting global brightness and color deviations.

\paragraph{MS-SSIM Loss.}
To better preserve structural consistency under varying illumination, we employ a multi-scale structural similarity (MS-SSIM)~\cite{MSSSIM} loss, defined as 
\begin{equation}
\label{eq:loss_msssim}
\mathcal{L}_{\text{MS-SSIM}}
= 1 - \prod_{m=1}^{M} \textmd{SSIM}^{(m)}\!\big(\hat{\mathbf{I}}, \mathbf{I}_{\text{GT}}\big),
\end{equation}
where $\textmd{SSIM}^{(m)}(\cdot, \cdot)$ is a structural similarity function computed at scale $m$, and $M$ is the set of all scales.
It captures contrast, luminance, and texture consistency between the prediction and ground truth at multiple spatial resolutions, which is particularly beneficial for low-light scenes where edges and fine details are difficult to recover.   

\paragraph{Perceptual Loss.}
Finally, to enhance perceptual realism and encourage faithful recovery of semantic structures, we incorporate a perceptual loss computed in a deep feature space. 
Letting $\Phi(\cdot)$ denote a fixed VGG-based feature extractor, the perceptual term is defined as 
\begin{equation}
\label{eq:loss_perc}
\mathcal{L}_{\text{Perc}}
= \frac{1}{N} 
\big\|  \Phi(\hat{\mathbf{I}}) - \Phi(\mathbf{I}_{\text{GT}})  \big\|^2_2.
\end{equation}
It is beneficial to compare the feature activations of the prediction with the ground truth.
This helps maintain natural textures and 
suppress color artifacts, two common failure points in low-light enhancement.

Together, the three losses form a complementary objective that encourages numerical accuracy, structural fidelity, and perceptual quality. 
In practice, we find that the hybrid loss significantly improves the visual consistency and robustness of Multinex compared to using a single loss component alone.  
We perform ablation study on loss terms in Section \ref{sec:loss_ablation}  and refer to results in Tab.~\ref{tab:loss_ablation}.

\subsection{Multinex Implementation} 

All experiments are conducted using the PyTorch framework~\cite{pytorch}.
Training data are augmented via random cropping, horizontal and vertical flipping, and random rotation.
Model parameters are optimized with the Adam optimizer~\cite{adam}, using a cosine annealing learning-rate schedule \cite{sgdr} that decays from $2\times10^{-4}$ to $1\times10^{-6}$.
Multinex is trained from scratch for 150K iterations with a batch size of 8 and patch size of $256\times256$, using the designated training splits of each dataset.

\subsection{Discussion on  GT-Mean}

 GT-Mean is a post-processing step used by some LLIE works   ~\cite{yan2025hvi,GLARE,hou2024global,LLFlow}  when evaluating their  approaches on small paired datasets, e.g., LOL-v1.
We do not perform GT-Mean in our assessment, as it removes the global brightness errors from the evaluation that actually is a core part of the LLIE performance.
In more detail, GT-Mean rescales the output of the enhanced image $\hat{\mathbf{I}}$ to match the mean grayscale of the ground truth image $\mathbf{I}_{\text{GT}}$, before computing PSNR/SSIM.
The rescaling is defined as $\hat{\mathbf{I}}_{\text{GT-Mean}} = q \hat{\mathbf{I}}$ with  $q =
\frac{\operatorname{mean}(\mathbf{I}_{\text{GT}})}
{\operatorname{mean}(\hat{\mathbf{I}})}$, and such a processing removes the global brightness errors.
We do not perform this rescaling,  as brightness correction is a core part of LLIE.
Enforcing matched luminance defeats the purpose of measuring enhancement accuracy.  
Consequently, GT-Mean can inflate the  performance metrics, for instance,  by several dB of PSNR on LOL-v1, e.g.,  CIDNet rises from {23.81}dB to {28.20}dB, while RetinexFormer from {25.15}dB to {27.14}dB. 
We report all the results \emph{without} GT-Mean to ensure that our evaluation reflects the true quality of LLIE enhancement.

\section{Additional Ablation Studies}
\label{app:ablation}

To complement Sec.~\ref{exp:abl}, we conduct additional ablation experiments  to further validate the Multinex design.
Unless otherwise stated, we use the same dataset and configuration as in Sec.~\ref{exp:abl}, and report the performance in terms of both the PSNR and SSIM metrics.

\begin{table}[t]
\centering
\setlength{\tabcolsep}{6pt}
\renewcommand{\arraystretch}{1.15}

\begin{tabular}{cccccc}
\toprule
$\mathcal{L}_\text{MSE}$ & $\mathcal{L}_\text{MS-SSIM}$  & $\mathcal{L}_\text{Perc}$ & PSNR$\uparrow$ & SSIM$\uparrow$ \\
\midrule
\checkmark & \xmark & \xmark & 22.31 & 0.815 \\
\xmark & \checkmark & \xmark & 19.04 & 0.820 \\
\xmark & \xmark & \checkmark & 18.80 & 0.793 \\
\checkmark & \checkmark & \xmark & 22.43 & 0.830 \\
\checkmark & \xmark & \checkmark & 22.80 & 0.821 \\
\xmark & \checkmark & \checkmark & 19.74 & 0.838 \\
\rowcolor{gray!12}
\checkmark & \checkmark & \checkmark & \textbf{23.19} & \textbf{0.843} \\
\bottomrule
\end{tabular}

\caption{Ablation of loss functions.}
\label{tab:loss_ablation}
\end{table}

\subsection{On Loss Function}
\label{sec:loss_ablation} 

Tab.~\ref{tab:loss_ablation} evaluates  contributions of the three individual loss terms and their combinations  to the final enhancement quality.  
The MSE loss on its own already provides a strong baseline, reaching around 22dB PSNR with a moderate SSIM of about 0.82.  
The MS-SSIM loss alone preserves quite well the structural similarity by offering a slightly higher SSIM, but it yields a noticeably lower PSNR, i.e., around 19dB.  
This indicates that MS-SSIM emphasizes contrast consistency over pixel-wise accuracy.  
The perceptual loss alone performs the worst in terms of distortion.
This is expected  since it optimizes high-level features rather than low-level fidelity.  
Pairwise combinations are able to improve the enhancement performance.
For instance, the combination of  MSE and  MS-SSIM slightly strengthens the structural consistency, while MSE and Perceptual together increase  brightness and color realism.  
A full combination of all the three losses yields the best results, reaching roughly 23dB PSNR with an SSIM of around 0.84.
This suggests that the pixel-level, structural, and perceptual cues are all important, while being complementary, for low-light enhancement.

\begin{table}[t]
\centering
\footnotesize
\setlength{\tabcolsep}{5.8pt}
\renewcommand{\arraystretch}{1.12}

\begin{tabular}{c c c c c c c}
\toprule
\multicolumn{4}{c}{\textbf{$\mathcal{S}_\mathcal{L}$ Components}} &
\#Ch & PSNR$\uparrow$ & SSIM$\uparrow$ \\
\cmidrule(lr){1-4}
$\mathbf{Y}_\text{Rec.709}$ & $\mathbf{Y}_\text{vmax}$ & $\mathbf{Y}_\text{lightness}$ & $\mathbf{Y}_{\text{L}_2}$ & & & \\
\midrule

\checkmark & \xmark & \xmark & \xmark & 1 & 18.80 & 0.753 \\
\xmark & \checkmark & \xmark & \xmark & 1 & 19.10 & 0.744 \\
\xmark & \xmark & \checkmark & \xmark & 1 & 19.23 & 0748 \\
\xmark & \xmark & \xmark & \checkmark & 1 & 19.05 & 0.757 \\
\midrule
\checkmark & \checkmark & \xmark & \xmark & 2 & 19.83 & 0.762 \\  
\checkmark & \xmark & \checkmark & \xmark & 2 & 20.12 & 0.776 \\  
\checkmark & \xmark & \xmark & \checkmark & 2 & 19.65 & 0.788 \\  
\xmark & \checkmark & \checkmark & \xmark & 2 & 20.84 & 0.795 \\  
\xmark & \checkmark & \xmark & \checkmark & 2 & 20.21 & 0.770 \\  
\xmark & \xmark & \checkmark & \checkmark & 2 & 20.55 & 0.782 \\  
\midrule
\checkmark & \checkmark & \checkmark & \xmark & 3 & 22.05 & 0.825 \\  
\checkmark & \checkmark & \xmark & \checkmark & 3 & 21.89 & 0.808 \\  
\checkmark & \xmark & \checkmark & \checkmark & 3 & 22.23 & 0.813 \\  
\xmark & \checkmark & \checkmark & \checkmark & 3 & 22.74 & 0.829 \\  
\midrule
\rowcolor{gray!12}
\checkmark & \checkmark & \checkmark & \checkmark & 4 & \textbf{23.19} & \textbf{0.843} \\

\bottomrule
\end{tabular}

\caption{Ablation studies on  feature maps of luminance guidance stack, supported by a complete Multinex architecture.}
\label{tab:illum_ablation_marks}
\end{table}

\subsection{On Luminance Guidance Stack}
Tab.~\ref{tab:illum_ablation_marks} analyzes the contribution of each individual illumination feature map  of $  \mathbf{Y}_\text{Rec.709}$, $  \mathbf{Y}_\text{vmax}$, $ \mathbf{Y}_\text{lightness}$, and $\mathbf{Y}_{\text{L}_2} $, also  their combinations.
When used alone, they provide   limited benefit, remaining in the range of 18-19dB PSNR with SSIM values just below 0.76.  
With pairwise combination, the performance consistently improves to the 20dB range, with the pair $(\mathbf{Y}_\text{vmax},\mathbf{Y}_\text{lightness})$ performing slightly better than the others.
This indicates that  exposure adjustment can be stabilized by  mixing physically grounded and perceptually aligned brightness cues.
The three-component combinations increase the performance further to around 22dB, showing that the different feature maps contribute useful complementary information, instead of being just equivalent variants of each other.  
The full combination of four maps leads to the best result, reaching roughly 23dB PSNR and 0.84 SSIM.
This confirms  that a multi-view luminance prior helps the Multinex illumination network $f_\mathcal{L}$ infer the required luminance adjustment more accurately.

\begin{table}[t]
\centering
\footnotesize
\setlength{\tabcolsep}{5.8pt}
\renewcommand{\arraystretch}{1.12}

\begin{tabular}{c c c c c c}
\toprule
\multicolumn{3}{c}{\textbf{$\mathcal{S}_\mathcal{R}$ Components}} &
\#Ch & PSNR$\uparrow$ & SSIM$\uparrow$ \\
\cmidrule(lr){1-3}
$[\mathbf{C_b,C_r}]$ & $[\mathbf{r,g}]$ & $\mathbf{S}$ & & & \\
\midrule

\checkmark & \xmark     & \xmark & 2 & 21.55 & 0.792 \\  
\xmark     & \checkmark & \xmark & 2 & 21.90 & 0.805 \\  
\xmark     & \xmark     & \checkmark & 1 & 20.62 & 0.779 \\  
\midrule
\checkmark & \checkmark & \xmark & 4 & 22.98 & 0.835 \\  
\checkmark & \xmark     & \checkmark & 3 & 22.12 & 0.819 \\  
\xmark     & \checkmark & \checkmark & 3 & 22.65 & 0.826 \\  
\midrule
\rowcolor{gray!12}
\checkmark & \checkmark & \checkmark & 5 & \textbf{23.19} & \textbf{0.843} \\

\bottomrule
\end{tabular}

\caption{Ablation studies on  feature maps of reflectance guidance stack, supported by a complete Multinex architecture.  }
\label{tab:chroma_ablation_marks}
\end{table}

\subsection{On Reflectance Guidance Stack}
Tab.~\ref{tab:chroma_ablation_marks} presents a similar ablation for feature maps of the reflectance guidance stack.  
The individual and combined contribution of the   chromaticity pair $[\mathbf{r}, \mathbf{g}]$,   YCbCr pair, and  saturation $\mathbf{S}$ are examined.
For individual contribution, the chromaticity pair $[\mathbf{r}, \mathbf{g}]$ performs the best, offering a PSNR around 22dB and SSIM slightly above 0.80,  likely because it provides illumination-invariant color ratios.  
The YCbCr pair offers slightly lower performance, while saturation alone is the weakest due to its sensitivity to noise. 
Pairwise combination  yields clear improvements.
For example, $[\mathbf{C}_b,\mathbf{C}_r]$ together with $[\mathbf{r},\mathbf{g}]$ achieves close to 23dB PSNR with the highest SSIM within the pairwise combination group.
Combinations involving saturation improve slightly less on SSIM, due to its noisier behavior.  
The complete three-way combination  yields the best performance with a  PSNR around 23dB and SSIM around mid-0.84, showing that the full  stack provides a balanced and complementary chroma representation for the Multinex reflectance network to learn effectively the reflectance adjustment.

\begin{table}[t]
\centering
\scriptsize
\setlength{\tabcolsep}{4pt}
\renewcommand{\arraystretch}{2.5}

\begin{tabular}{l l c c}
\toprule
\textbf{Placement} &
\textbf{Formulation} $f(\mathcal{S})=\mathrm{Conv}_{1\times1} \circ\dots$ &
PSNR$\uparrow$ & SSIM$\uparrow$ \\
\midrule

Before  &
$ \mathrm{FB}^{2T}\Big(\mathrm{Conv}_{1\times1}(\mathcal{S}) \odot \mathrm{CWA}(\mathcal{S})\Big) $
& 22.78 & 0.831 \\

\rowcolor{gray!12}
Between  &
$ \mathrm{FB}^T \Big( \mathrm{CWA}(\mathcal{S}) \odot \mathrm{FB}^T \circ \mathrm{Conv}_{1\times1}(\mathcal{S})  \Big)$
& \textbf{23.19} & \textbf{0.843} \\

After  &
$  \mathrm{CWA}(\mathcal{S}) \odot \mathrm{FB}^{2T}\circ \mathrm{Conv}_{1\times1}(\mathcal{S})   $
& 22.41 & 0.823 \\

\bottomrule
\end{tabular}

\caption{Ablation on where to place CWA in fusion networks.}
\vspace{-7pt}
\label{tab:cwa_ablation_full}
\end{table}

\subsection{On CWA Placement }

In this section, we examine the effect of where to place   CWA   within the fusion networks, by experimenting with three ways to insert the CWA mechanism  as listed in  Tab.~\ref{tab:cwa_ablation_full}, where the ``between" setting corresponds to the proposed design in Eq. (\ref{eq:fusion}).
In the other two settings of ``before" and ``after", we also increase  the layer depth  by applying the FB modules $2T$ times instead of $T$.
By applying CWA early, i.e., before the main fusion blocks, we can obtain good result  of roughly 23dB PSNR and 0.83 SSIM. 
By placing CWA after  deeper FBs, we obtain  slightly worse performance, indicating that a late attention is less effective when the features have already been heavily mixed.  
Our adopted design inserts CWA between the projection and the stacked FBs, which yields the best performance of a similar 23dB PSNR while higher SSIM  close to 0.84.  
Overall, despite having  deeper layers, the other two ways of placing CWA decrease the model performance.
This suggests that the proposed approach of  weighting the analytic feature maps early, followed by a lightweight spatial refinement, is an effective strategy for leveraging multi-view representation priors under tight parameter constraints.

\section{More Qualitative Examples}
\label{app:examples}

\begin{figure*}
    \centering
    \includegraphics[width=1.0\linewidth, trim=0 0 {4.5\linewidth} 0, clip]{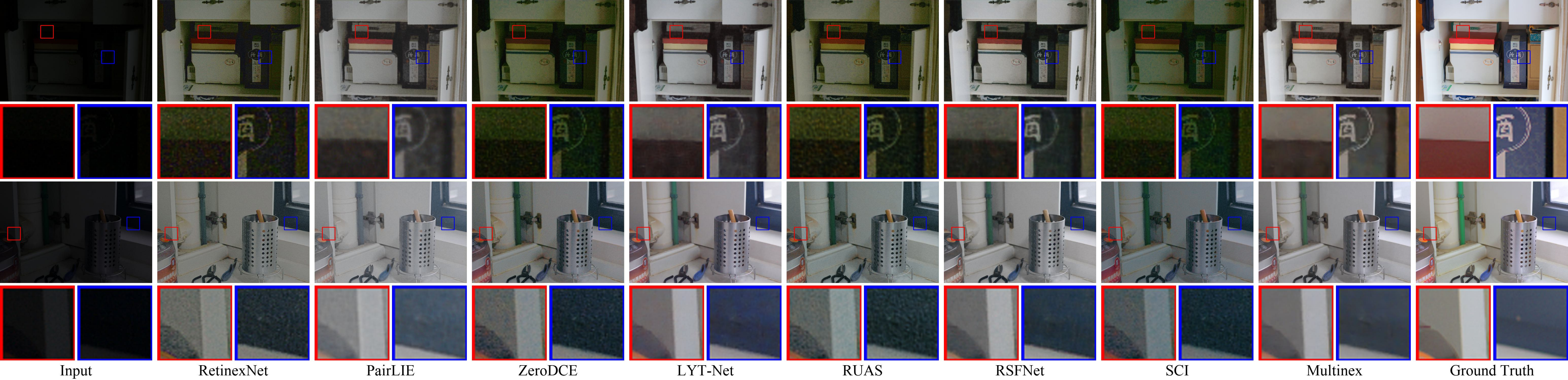}
    \includegraphics[width=1.0\linewidth, trim={4.5\linewidth} 0 0 0, clip]{fig_suppl/suppl_LOL_lightweight_1.pdf}
\caption{Qualitative comparison on reference dataset LOL-v1~\cite{RetinexNet} between Multinex and state-of-the-art lightweight and micro scale models RetinexNet~\cite{RetinexNet}, PairLIE~\cite{pairlie}, ZeroDCE~\cite{zero_dce}, LYT-Net~\cite{brateanu2025lyt}, RUAS~\cite{RUAS}, RSF~\cite{RSFNet}, SCI~\cite{SCI}.
}
    \label{fig:suppl_LOL_lightweight_1}
\end{figure*}

\begin{figure*}
    \centering
    \includegraphics[width=1.0\linewidth, trim=0 0 {4.5\linewidth} 0, clip]{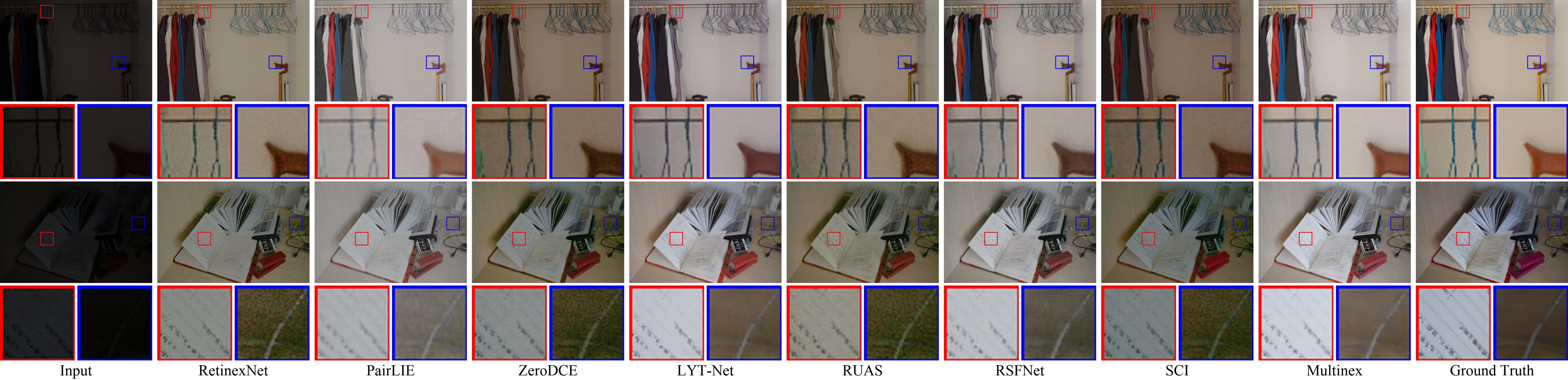}
    \includegraphics[width=1.0\linewidth, trim={4.5\linewidth} 0 0 0, clip]{fig_suppl/suppl_LOL_lightweight_2.pdf}
    \caption{Qualitative comparison on reference dataset LOL-v1~\cite{RetinexNet} between Multinex and state-of-the-art lightweight and micro scale models RetinexNet~\cite{RetinexNet}, PairLIE~\cite{pairlie}, ZeroDCE~\cite{zero_dce}, LYT-Net~\cite{brateanu2025lyt}, RUAS~\cite{RUAS}, RSF~\cite{RSFNet}, SCI~\cite{SCI}. }
    \label{fig:suppl_LOL_lightweight_2}
\end{figure*}

\begin{figure*}
    \centering
    \includegraphics[width=1.0\linewidth, trim=0 55pt 0 0, clip]{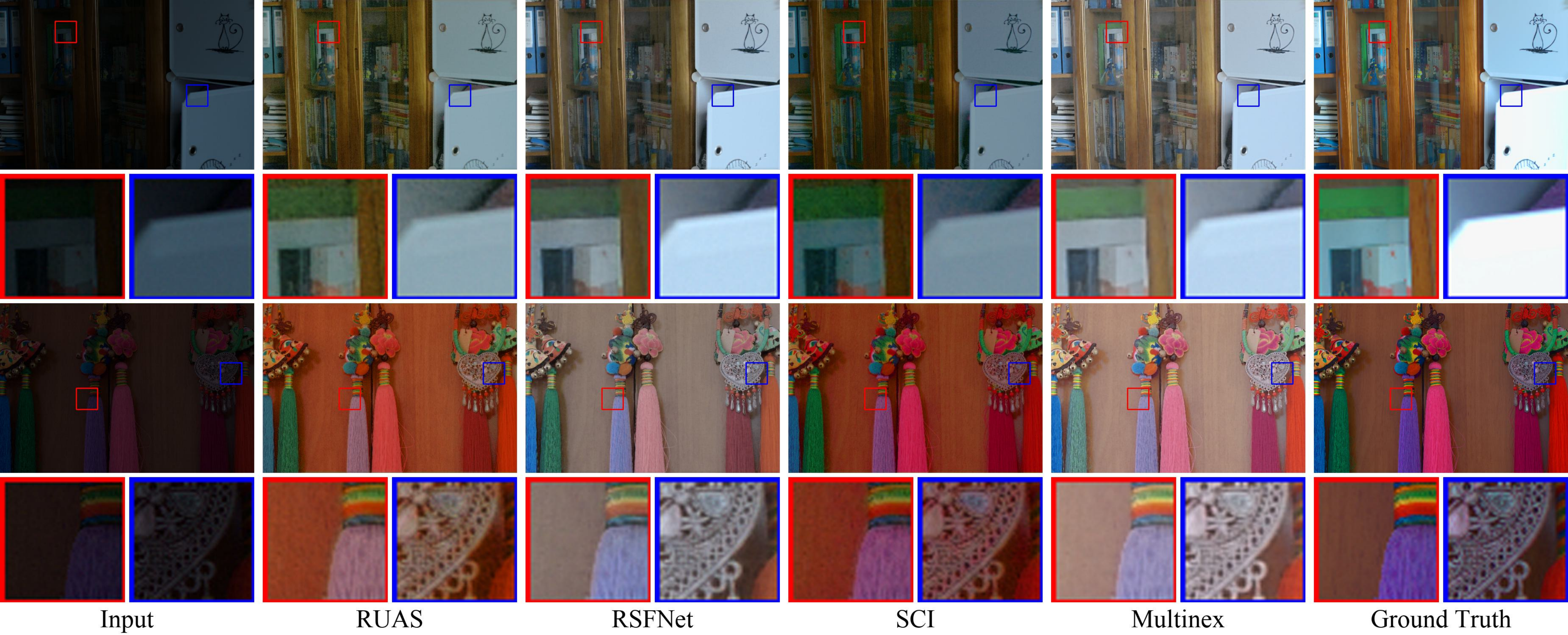}
    \includegraphics[width=1.0\linewidth, trim=0 55pt 0 0, clip]{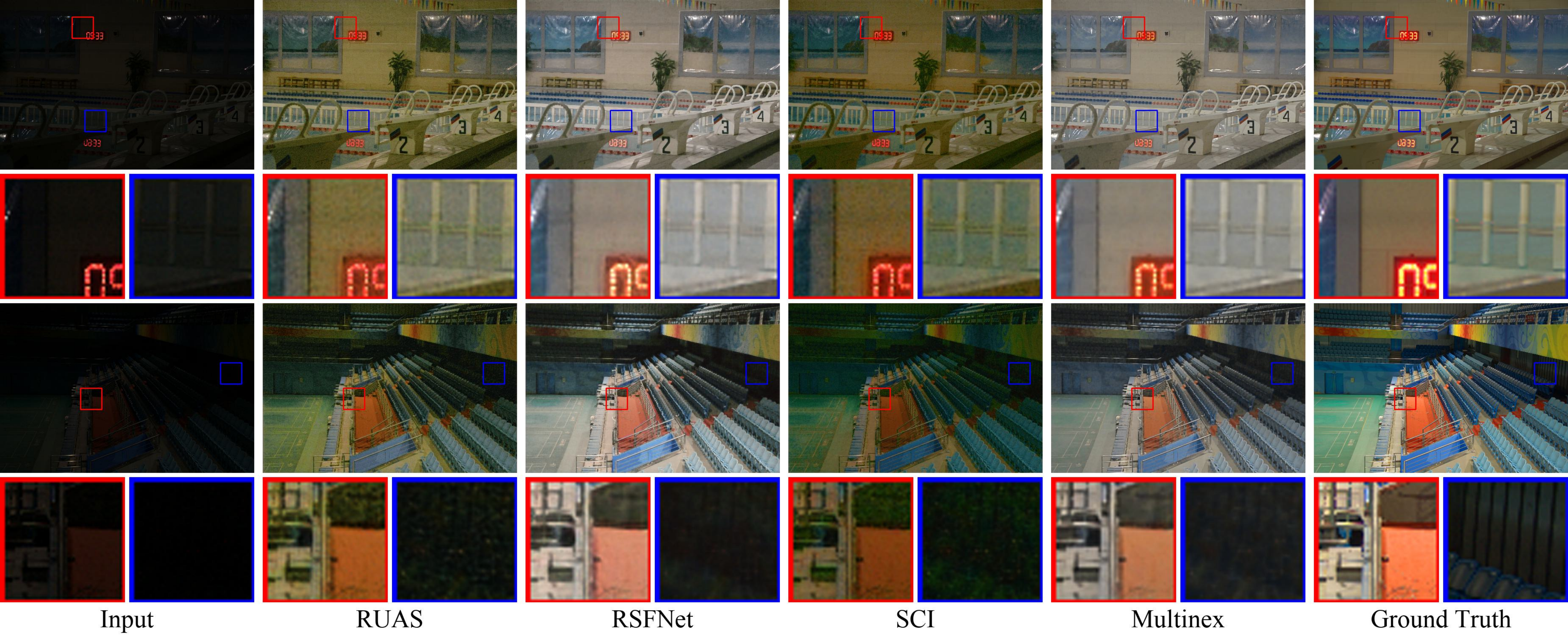}
    \includegraphics[width=1.0\linewidth, trim=0 0 0 0, clip]{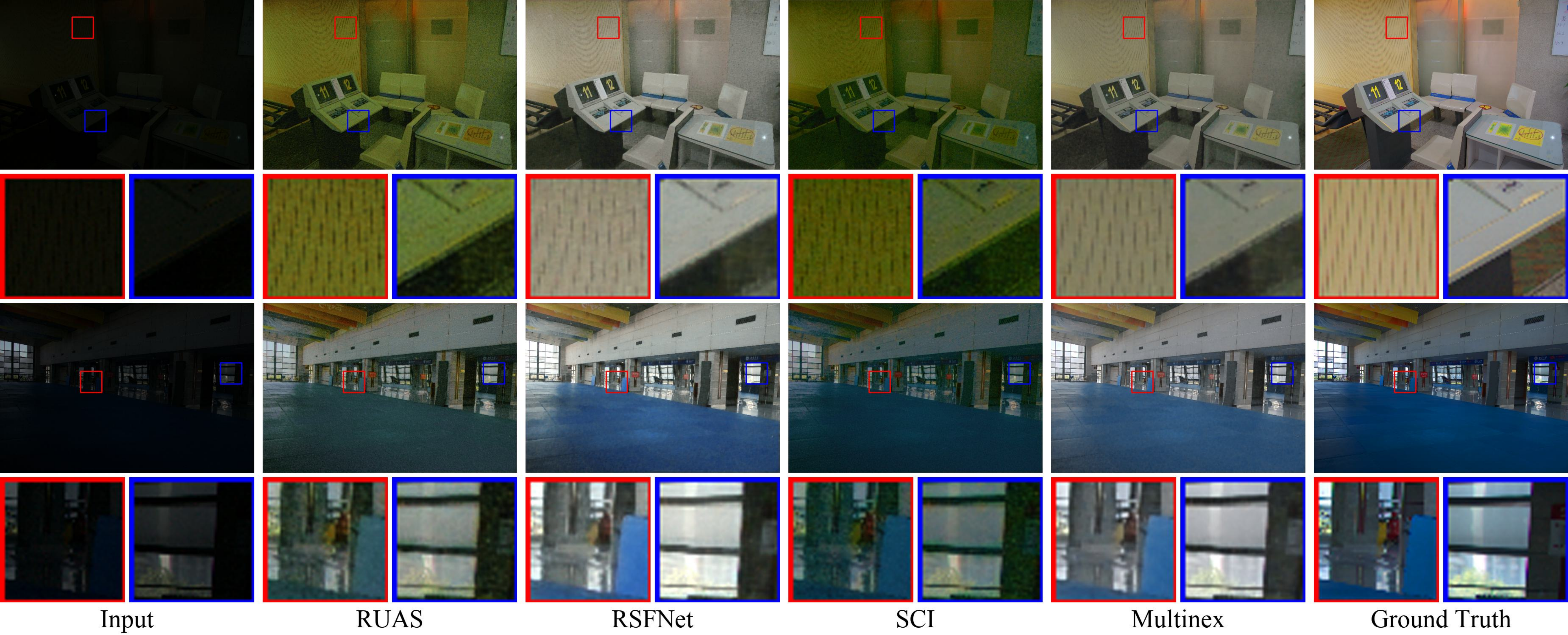}

    \caption{Qualitative comparison on reference dataset LOL-v1~\cite{RetinexNet} between Multinex-Nano and state-of-the-art micro scale models RUAS~\cite{RUAS}, RSF~\cite{RSFNet}, SCI~\cite{SCI}. }
    \label{fig:suppl_LOL_micro}
\end{figure*}

\begin{figure*}
    \centering
    \includegraphics[width=1.0\linewidth, trim=0 55pt 0 0, clip]{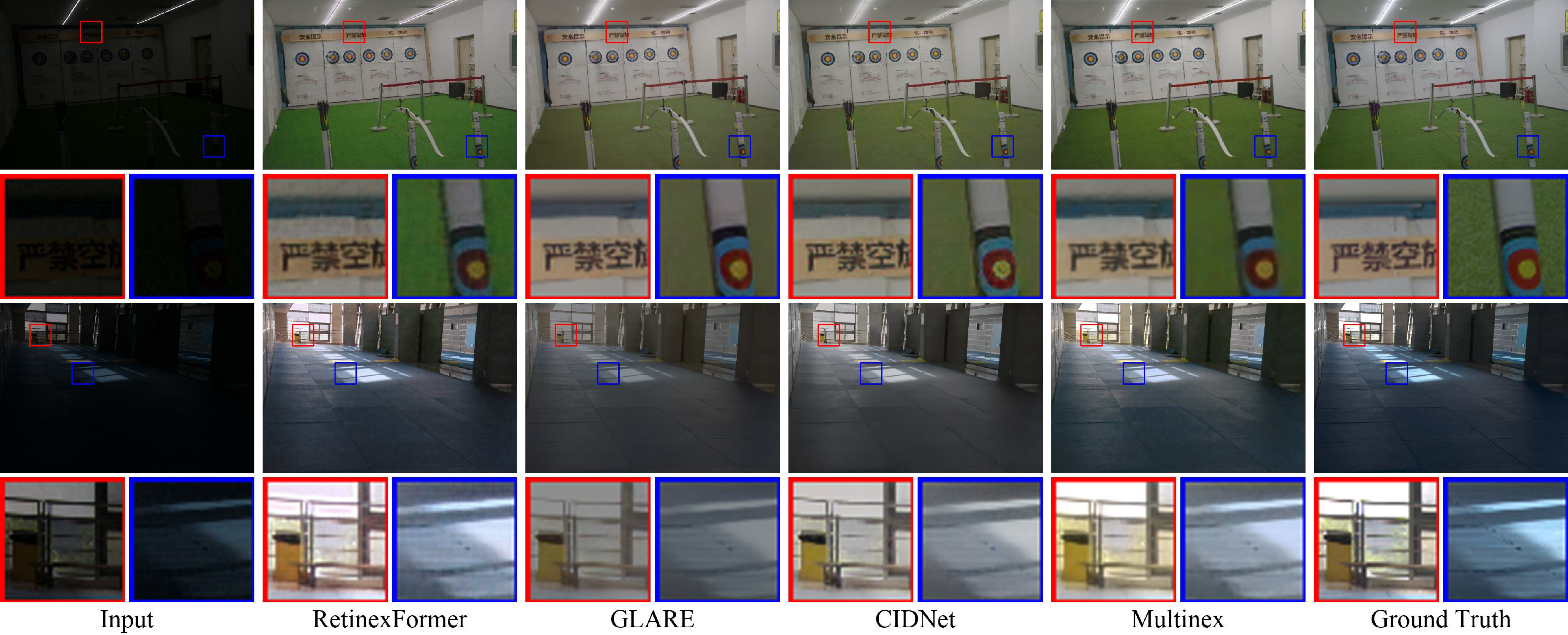}
    \includegraphics[width=1.0\linewidth, trim=0 55pt 0 0, clip]{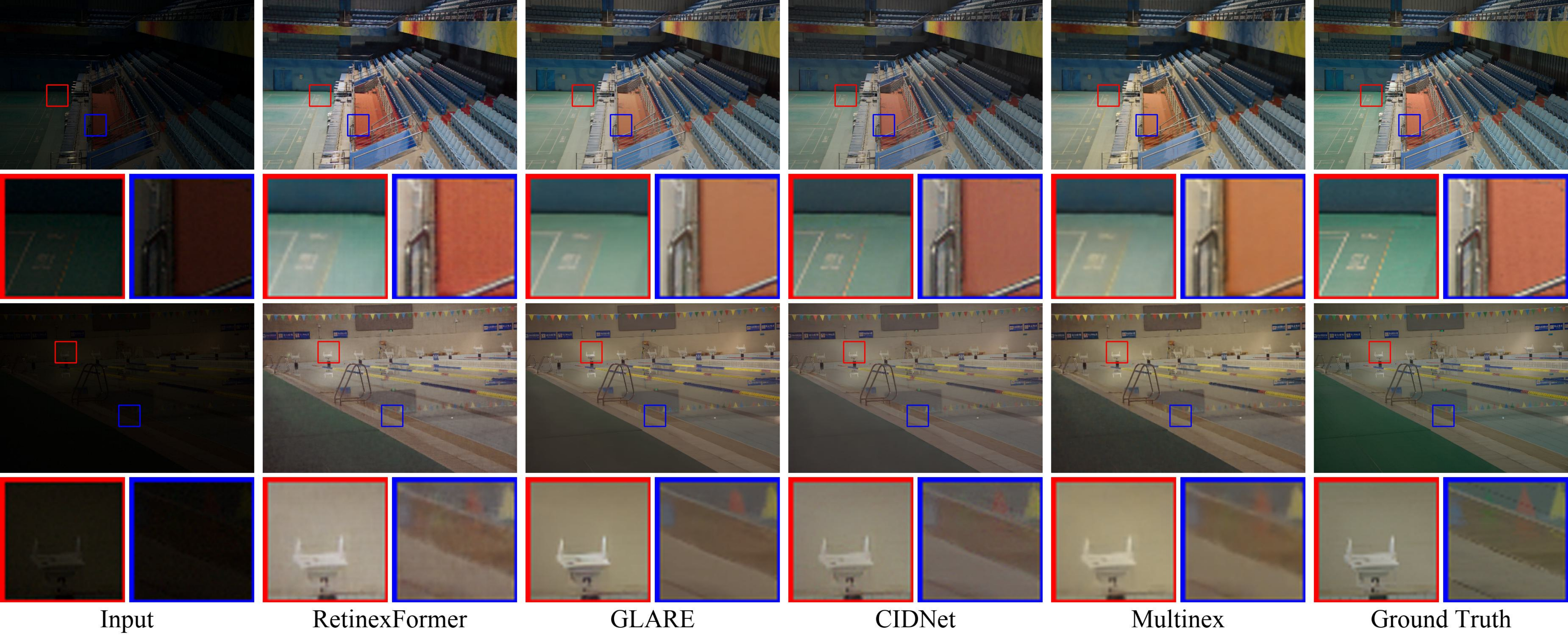}
    \includegraphics[width=1.0\linewidth]{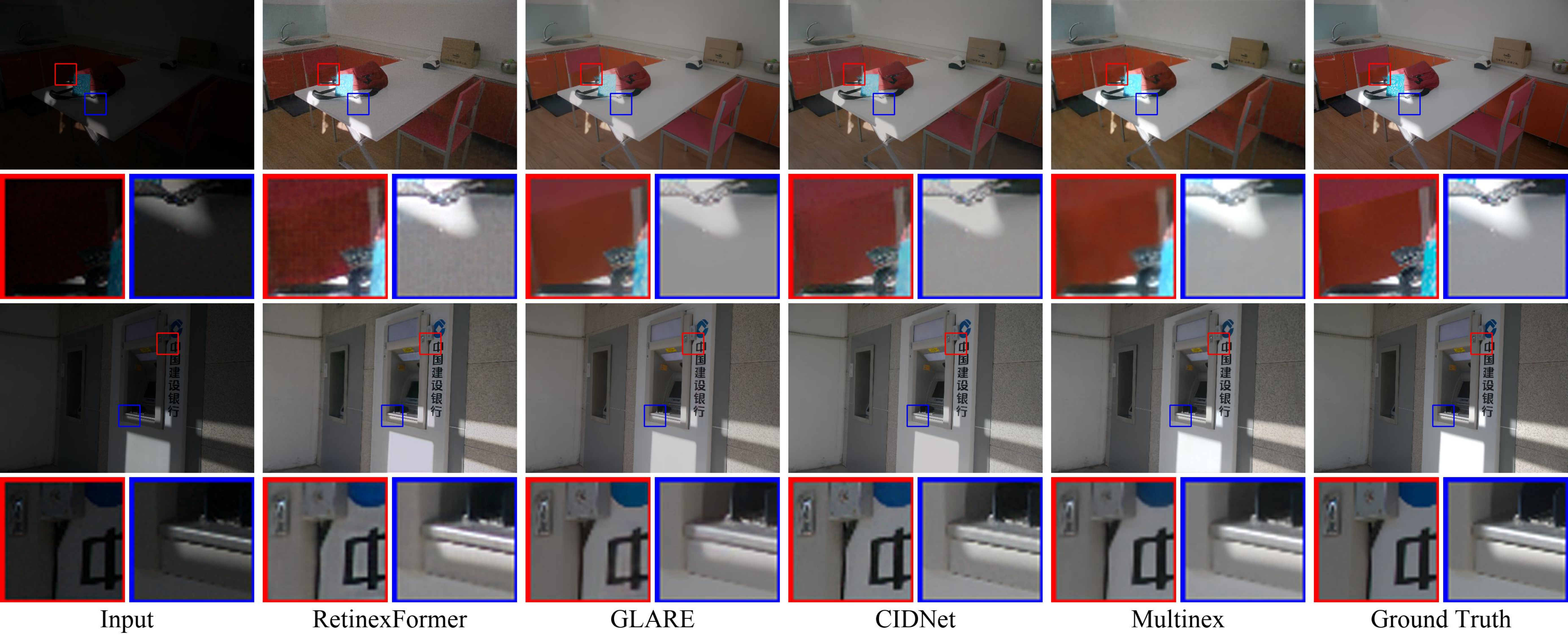}
    \caption{Qualitative comparison on reference dataset LOL-v2-real~\cite{Sparse} between Multinex and state-of-the-art mid-sized (1-10 M param.) models RetinexFormer~\cite{Retinexformer} and CIDNet~\cite{yan2025hvi}, and heavy ($>$10 M param.) model GLARE~\cite{GLARE}. }
    \label{fig:suppl_LOL_mid_heavy}
\end{figure*}

\begin{figure*}
    \centering
    \includegraphics[width=0.8\linewidth, trim=0 0 {0.81\linewidth} 0, clip]{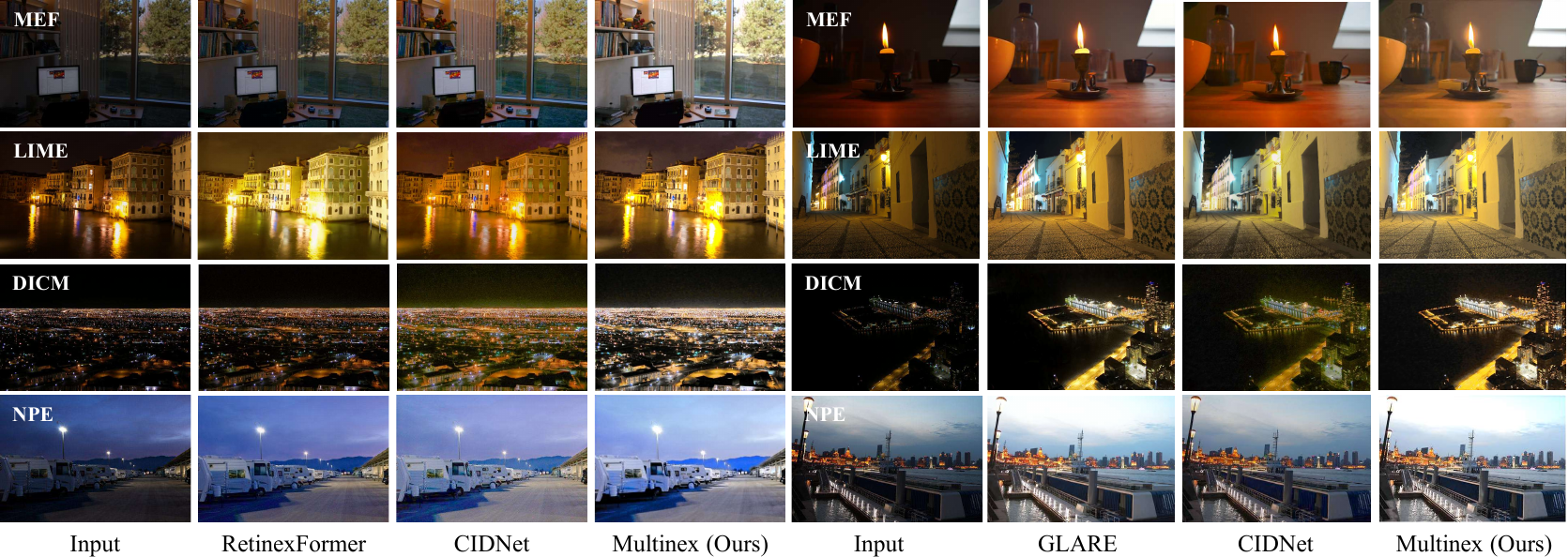}
    \includegraphics[width=0.8\linewidth, trim={0.82\linewidth} 0 0 0, clip]{fig/noref_horizontal_big_cropped.pdf}
    \caption{Qualitative comparison on no-reference datasets MEF~\cite{MEF}, LIME~\cite{LIME}, DICM~\cite{DICM}, NPE~\cite{NPE} between Multinex and state-of-the-art mid-sized (1-10M param.) models RetinexFormer~\cite{Retinexformer} and CIDNet~\cite{yan2025hvi} and heavy-weight ($>$10M param.) GLARE~\cite{GLARE}. }
    \label{fig:noref_qualitative}
\end{figure*}

\begin{figure*}
    \centering
    \includegraphics[width=0.88\linewidth, trim=0 0 0 0, clip]{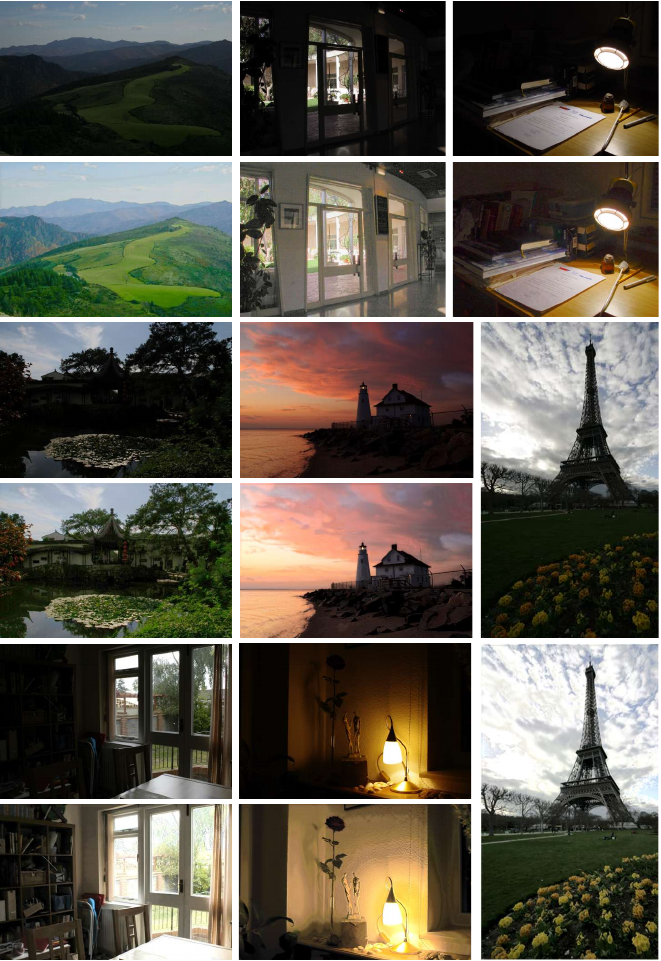}
    \caption{Additional results on no-reference dataset MEF~\cite{MEF}. For corresponding images, top is input, and bottom is Multinex output.}
    \label{fig:suppl_mef_extra}
\end{figure*}

\begin{figure*}
    \centering
    \includegraphics[width=0.9\linewidth, trim=0 0 0 0, clip]{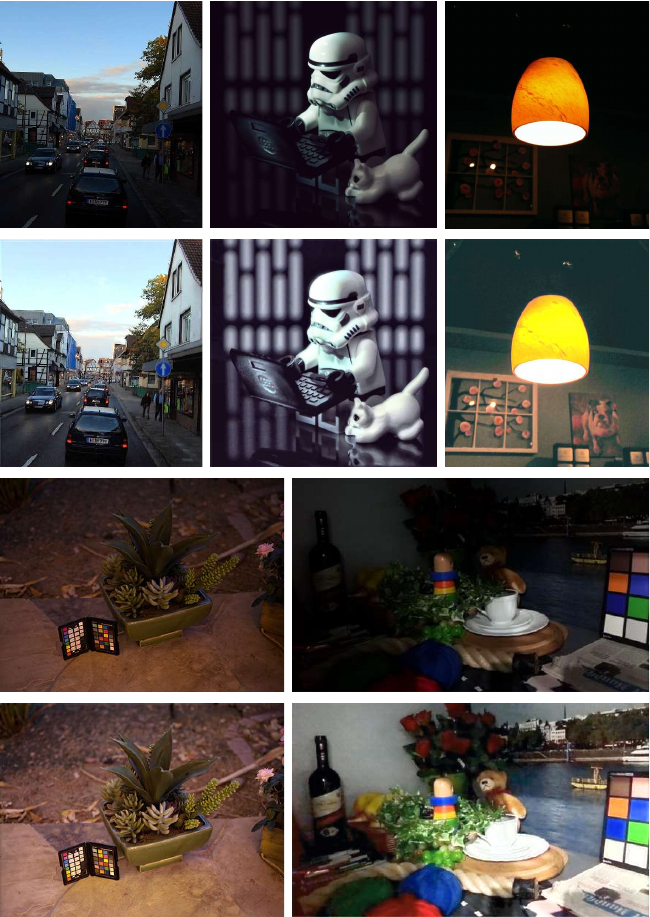}
    \caption{Additional results on no-reference dataset LIME~\cite{LIME}. For corresponding images, top is input, and bottom is Multinex output.}
    \label{fig:suppl_lime_extra}
\end{figure*}

\begin{figure*}
    \centering
    \includegraphics[width=0.85\linewidth, trim=0 0 0 0, clip]{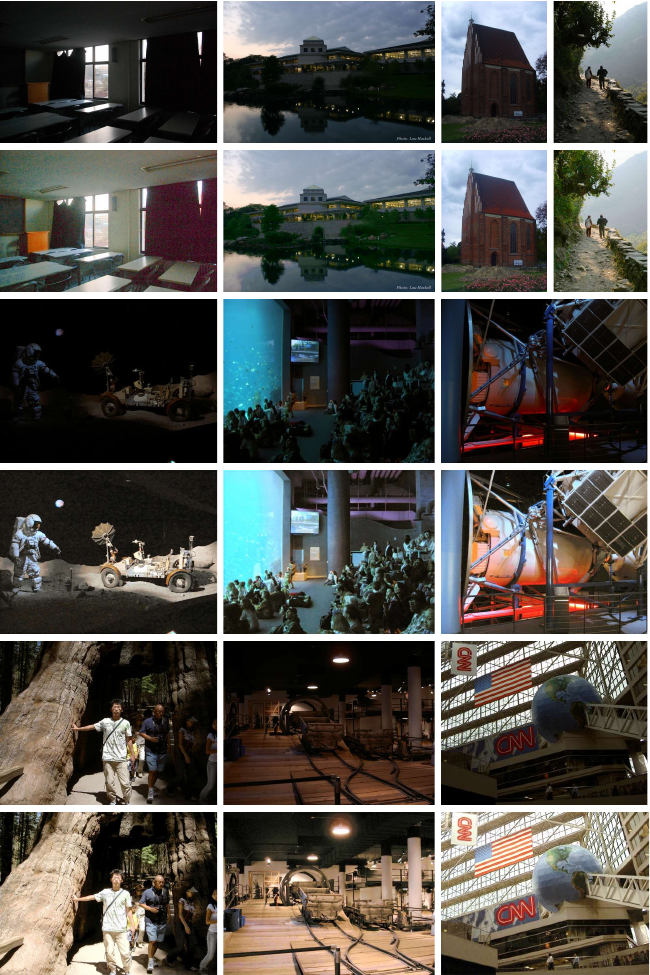}
    \caption{Additional results on no-reference dataset DICM~\cite{DICM}. For corresponding images, top is input, and bottom is Multinex output.}
    \label{fig:suppl_dicm_extra}
\end{figure*}

\begin{figure*}
    \centering
    \includegraphics[width=0.85\linewidth, trim=0 0 0 0, clip]{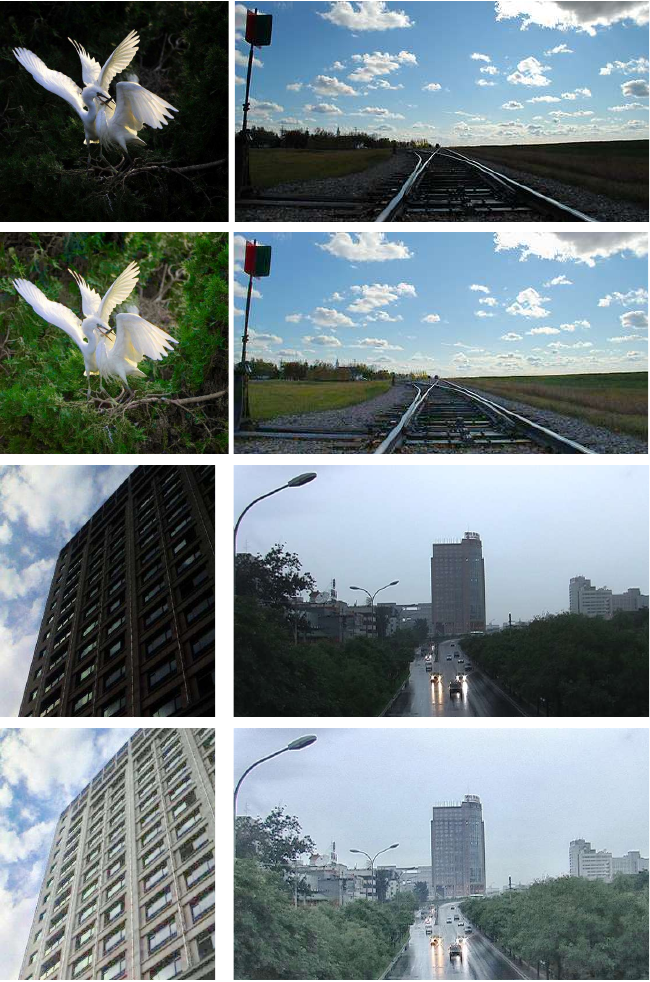}
    \caption{Additional results on no-reference dataset NPE~\cite{NPE}. For corresponding images, top is input, and bottom is Multinex output.}
    \label{fig:suppl_npe_extra}
\end{figure*}

We provide additional qualitative comparisons and results on various datasets.
It can be seen from Figs.~\ref{fig:suppl_LOL_lightweight_1} and \ref{fig:suppl_LOL_lightweight_2} that Multinex achieves better color fidelity and detail recovery as compared to other previous lightweight and micro approaches, while also attain  overall better brightness recovery.
%
In Fig.~\ref{fig:suppl_LOL_micro}, Multinex-Nano shows significantly better level of detail and illumination correction compared to prior micro-sized approaches.
In Fig.~\ref{fig:suppl_LOL_mid_heavy}, Multinex shows strong performance, approaching (sometimes outperforming) heavier models. 
While level of detail is lower in some cases, illumination and color correction are stronger, despite having a significantly smaller model size.
Fig. ~\ref{fig:noref_qualitative} shows that Multinex exhibits greater color fidelity with stronger brightness improvements. 
However, it produces less-detailed outputs as compared to GLARE, due to the massive parameter scale difference.
Finally, we demonstrate more examples of low-light images and their enhanced images predicted by Multinex in Figs.~\ref{fig:suppl_mef_extra}~\ref{fig:suppl_lime_extra}~\ref{fig:suppl_dicm_extra}~\ref{fig:suppl_npe_extra}.
%

\begin{figure*}
    \centering
    \includegraphics[width=0.8\linewidth]{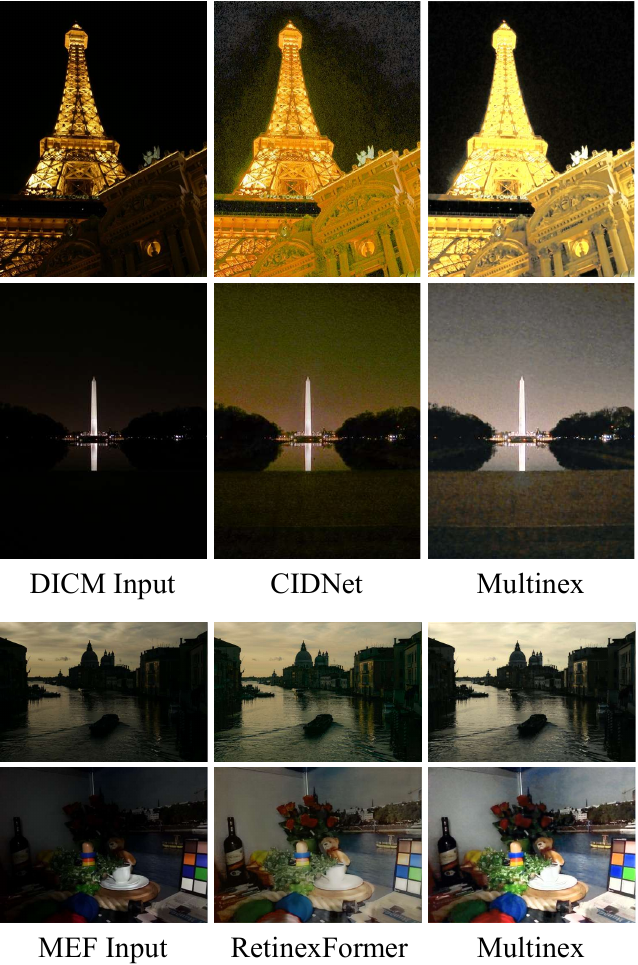}
    \caption{A few challenging cases from DICM and MEF datasets.}
    \label{fig:suppl_challenging}
\end{figure*}

For a matter of interest, we also show in Fig.~\ref{fig:suppl_challenging}  a few very challenging low-light images.
It can be seen that Multinex produces light distortions and noise on the DICM examples, and exhibits color loss on the MEF examples due to over-exposure.
This is primarily due to its low parameter count which hinders detail reconstruction and color balancing in challenging scenarios.
However, it can be seen from Fig~\ref{fig:suppl_challenging} that images enhanced by those mid-sized approaches like RetinexFormer~\cite{Retinexformer} and CIDNet~\cite{yan2025hvi} also show artifacts.
%

%

\clearpage

\clearpage

\clearpage

\newlength{\halfwidth}
\setlength{\halfwidth}{0.5\linewidth}

\clearpage

\clearpage

\clearpage

\clearpage

\clearpage

\clearpage






\end{document}